\documentclass[letterpaper]{article} % DO NOT CHANGE THIS
\usepackage{aaai25}  % DO NOT CHANGE THIS
\usepackage{times}  % DO NOT CHANGE THIS
\usepackage{helvet}  % DO NOT CHANGE THIS
\usepackage{courier}  % DO NOT CHANGE THIS
\usepackage[hyphens]{url}  % DO NOT CHANGE THIS
\usepackage{graphicx} % DO NOT CHANGE THIS
\usepackage{natbib}  % DO NOT CHANGE THIS AND DO NOT ADD ANY OPTIONS TO IT
\usepackage{caption} % DO NOT CHANGE THIS AND DO NOT ADD ANY OPTIONS TO IT
\usepackage{algorithm}
\usepackage{algorithmic}

\usepackage{newfloat}
\usepackage{listings}
\DeclareCaptionStyle{ruled}{labelfont=normalfont,labelsep=colon,strut=off} % DO NOT CHANGE THIS
\floatstyle{ruled}
\newfloat{listing}{tb}{lst}{}
\floatname{listing}{Listing}
\usepackage{amssymb}
\usepackage{adjustbox}
\usepackage{booktabs}
\usepackage{amsmath}
\usepackage{subfig}
\usepackage{amsfonts}
\newtheorem{theorem}{Theorem}
\newtheorem{proof}{Proof}
\newtheorem{assumption}{Assumption}
\newtheorem{lemma}{Lemma}

\newtheorem{remark}{Remark}
\usepackage{multirow}
\usepackage{xcolor}

\title{In-Dataset Trajectory Return Regularization for Offline Preference-based Reinforcement Learning}
\author{
    Songjun Tu\textsuperscript{\rm 1,2,3}, Jingbo Sun\textsuperscript{\rm 1,2,3}, Qichao Zhang\textsuperscript{\rm 1,3}\thanks{Corresponding author. \\ Code Page: {https://github.com/TU2021/RL-SaLLM-F}}, Yaocheng Zhang\textsuperscript{\rm 1,3}, \\
    Jia Liu\textsuperscript{\rm 4}, Ke Chen\textsuperscript{\rm 2}, Dongbin Zhao\textsuperscript{\rm 1,2,3} 
}
\affiliations{
    \textsuperscript{\rm 1}State Key Laboratory of Multimodal Artificial Intelligence Systems, CASIA, Beijing, China\\
    \textsuperscript{\rm 2}Peng Cheng Laboratory, Shenzhen, China \\
    \textsuperscript{\rm 3}School of Artificial Intelligence, University of Chinese Academy of Sciences, Beijing, China \\
    \textsuperscript{\rm 4}School of Mathematics and Statistics, Xi'an Jiaotong University, Xi'an, China \\
    \{tusongjun2023,zhangqichao2014\}@ia.ac.cn
}

\usepackage{bibentry}
\begin{document}

\maketitle

\begin{abstract}
Offline preference-based reinforcement learning (PbRL) typically operates in two phases: first, use human preferences to learn a reward model and annotate rewards for a reward-free offline dataset; second, learn a policy by optimizing the learned reward via offline RL.
However, accurately modeling step-wise rewards from trajectory-level preference feedback presents inherent challenges. 
The reward bias introduced, particularly the overestimation of predicted rewards, leads to optimistic trajectory stitching, which undermines the pessimism mechanism critical to the offline RL phase.
To address this challenge, we propose In-Dataset Trajectory Return Regularization (DTR) for offline PbRL, which leverages conditional sequence modeling to mitigate the risk of learning inaccurate trajectory stitching under reward bias.
Specifically, DTR employs Decision Transformer and TD-Learning to strike a balance between maintaining fidelity to the behavior policy with high in-dataset trajectory returns and selecting optimal actions based on high reward labels.
Additionally, we introduce an ensemble normalization technique that effectively integrates multiple reward models, balancing the trade-off between reward differentiation and accuracy. 
Empirical evaluations on various benchmarks demonstrate the superiority of DTR over other state-of-the-art baselines.

\end{abstract}

%     \link{Code}{https://github.com/TU2021/DTR}
% %     \link{Datasets}{https://aaai.org/example/datasets}
%     % \link{Extended version}{https://arxiv.org/html/2412.09104}

\section{Introduction}
Designing complex artificial rewards in reinforcement learning (RL) is challenging and time-consuming \cite{reward22,wang2024prototypical}. 
Preference-based reinforcement learning (PbRL) addresses this by leveraging human feedback to guide policies, demonstrating success in aligning large language models \cite{rlhf22} and robot control \cite{rune22}. 
Recently, considering the growing utilization of offline data in aiding policy optimization via offline RL \cite{fang2022offline,chen2024boosting}, offline PbRL \cite{oprl23} has gained attention. 
This approach involves training a reward model with limited human feedback, annotating rewards for a reward-free offline dataset, and applying offline RL  to learn policies.

Despite significant advancements in offline PbRL, learning an accurate step-wise reward model from trajectory-wise preference feedback remains inherently challenging due to limited feedback data \cite{ftb24}, credit assignment \cite{pt23} and neural network approximation errors \cite{theory23}. 
The introduced reward bias adds potential brittleness to the pipeline, leading to suboptimal performance \cite{yu2022leverage, hu2023provable}.
To mitigate this issue, some studies have aimed to enhance the robustness of the reward model \cite{oprl23, hpl24}, yet ignoring the potential influence of reward bias in offline RL.  
Alternatively, approaches that bypass reward modeling and directly optimize policy using preference \cite{dppo23, cpl24} struggle to achieve out-of-distribution (OOD) generalization, limiting their ability to outperform the dataset \cite{xu2024dpo}.

For the policy learning, most of offline PbRL methods \cite{rlhf17,uni-rlhf,hpl24} apply the learned reward function directly to downstream TD-Learning (TDL) based offline RL algorithms, such as CQL \cite{cql20} and IQL \cite{iql22}. 
However, these TDL-based methods do not account for potential bias in the predicted rewards.
The introduced reward bias, especially overestimated rewards, can lead to optimistic trajectory stitching and undermine the pessimism towards OOD state-action pairs in offline TDL algorithms \cite{yu2022leverage}.
To minimize the impact of reward bias, 
it is necessary to consider being pessimistic about overestimated rewards during the offline policy learning phase \cite{provable24}.
Apart from TDL-based offline algorithms, another methodology, conditional sequence modeling (CSM) \cite{rvs22} such as Decision Transformer (DT) \cite{dt21}, has not yet been investigated in offline PbRL.
This type of imitation-based approach learns a maximum return policy with in-dataset trajectories by assigning appropriate trajectory reweighting.
Compared to TDL, which extracts policy based on value functions, CSM extracts policy conditioned on in-dataset trajectory returns, thereby potentially {constraints its  trajectory stitching capability \cite{qdt23}}. 
Although a limited trajectory stitching ability may not be anticipated for offline RL with  ground-truth (GT) rewards, this limitation can be considered as a pessimistic safeguard to alleviate the inaccurate trajectory stitching due to incorrect reward labels {and maintain fidelity to the behavior policy with high in-dataset trajectory returns in offline PbRL.}
These properties trigger our further thought: 

\textit{Is it possible to leverage the trajectory return-based CSM to mitigate inaccurate stitching of TDL for offline PbRL?}

Building upon these insights, we propose 
{In-\textbf{D}ataset \textbf{T}rajectory Return  \textbf{R}egularization \textbf{(DTR)} for offline PbRL},
which integrates CSM and TDL to achieve the offline RL policy based on in-dataset trajectory returns and annotated step-wise preference rewards. Specifically,
DTR consists of the following three core components:
(1) a DT \cite{dt21} structure based policy is employed to associate the return-to-go (RTG) token with in-dataset individual trajectories, aintaining fidelity to the behavior policy with high trajectory-wise returns;
(2) a TDL module aims to utilize $Q$ function to select optimal actions with high step-wise rewards and balance the limited trajectory stitching ability of DT; and
(3) an ensemble normalization that effectively integrates multiple reward models to balance reward differentiation and inaccuracy.

By combining CSM and TDL dynamically, DTR mitigates the potential risk of reward bias in offline PbRL, leading to enhanced performance.
Our contributions are summarized below:
\begin{itemize}
    \item We propose DTR, a valid integration of CSM with DT-based regularization to address the impact of TD-learning stitching caused by reward bias in offline PbRL.

    \item We introduce a dynamic coefficient in the policy loss to balance the conservatism and exploration, and an ensemble normalization method in reward labeling to balance reward differentiation and inaccuracy.
    
    \item We prove that extracting policies from in-dataset trajectories leads to provable suboptimal bounds, resulting in enhanced performance in offline PbRL. 
    
    \item Our experiments on public datasets demonstrate the superior performance and significant potential of DTR.
\end{itemize}

\section{Related Works}
\subsubsection{Offline PbRL.} 
To enhance the performance of offline PbRL, some works emphasize the importance of improving the robustness of reward model, such as improving the credit assignment of returns \cite{non-markov22,pt23,prior24}, utilizing data augmentation to augment the generalization of reward model \cite{qpa24}. 
Other approaches modify the Bradley-Terry model to optimize preferences directly and avoid reward modeling, such as DPPO \cite{dppo23} and CPL \cite{cpl24}.  
{A related SOTA work is FTB \cite{ftb24}, which optimizes policies based on diffusion model and augmented trajectories without TDL.
In contrast, our approach balances trajectory-wise DT and step-wise TD3 \cite{td3_18} to mitigate the risk of reward bias and leads to enhanced performance.}

In theory, \cite{theory23} proves that the widely used maximum likelihood estimator (MLE) converges under the Bradley-Terry model in offline PbRL with the restriction to linear reward function. \cite{hu2023provable} stresses that pessimism about overestimated rewards should be considered in offline RL. 
\cite{provable24} relaxes the assumption of linear reward and provides theoretical guarantees to general function approximation.
Based on these results, we further provide theoretical suboptimal bound guarantees for offline PbRL under the estimated value function.

\subsubsection{Improving the Stitching Ability of CSM.} 
Due to the transformer architecture and the paradigm of supervised learning, CSM has limited trajectory stitching ability.
{Therefore, some works aggregate the input of the transformer-based policy or modify its structure to adapt to the characteristics of multimodal trajectories \cite{starformer22,gcpc24,dc24}.}
The alternative approach leverages $Q$ value to enhance trajectory stitching capability such as CGDT \cite{cgdt22}, QDT \cite{qdt23}, Q-Transformer \cite{qtransformer23} and recent QT \cite{qt24} for offline RL with GT rewards.

\begin{figure*}[h]
 \centering
 \includegraphics[width=\linewidth]{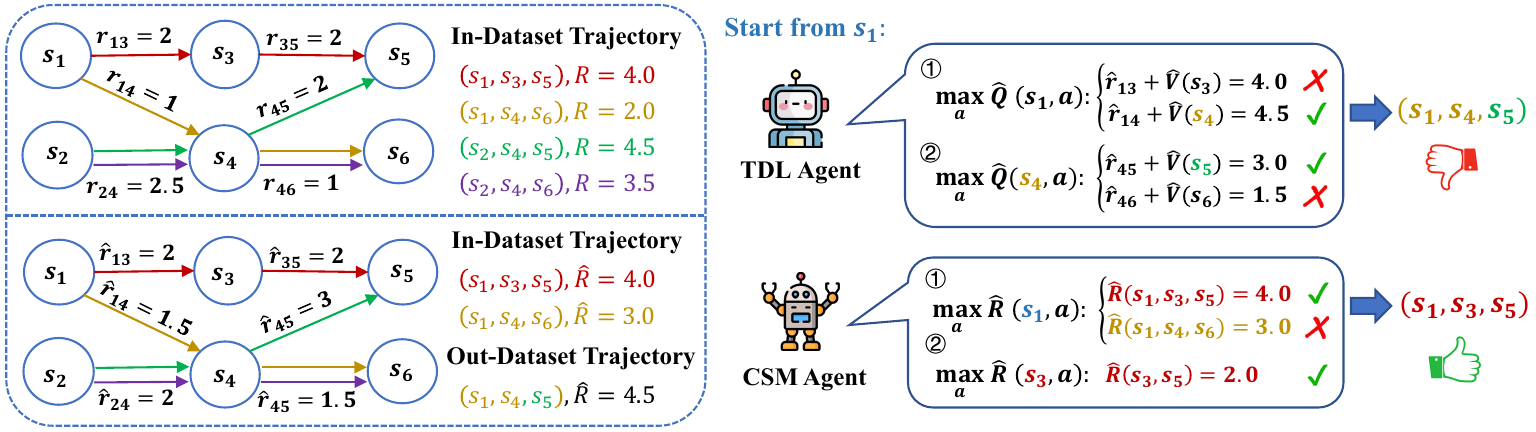}
 \caption{A deterministic MDP that illustrates the potential failure mode of offline PbRL.  
 Since the state transition is deterministic, we represent the trajectory as a sequence of states.
 The four colored lines represent four trajectories in the preference dataset $\mathcal{D}_{pref}$. 
 The reward $r$ indicates the GT reward, and $\hat{r}$ indicates the estimated reward from preference. {Starting at state $s_1$, inaccurate estimated rewards result in a greater return of the stitched trajectory $(s_1,s_4,s_5)$ than the in-dataset trajectory ${(s_1,s_3,s_5)}$ for TDL agent,} which is factually incompatible. { In contrast, CSM agent gets correct actions that align closely with the behavior policy conditioned on high in-dataset trajectory return.} }
 \label{fig:mdp}
\end{figure*}

\section{Preliminaries}

\subsection{Learning Rewards From Human Feedback}

Following previous studies \cite{pebble21,pt23}, we consider trajectories of length $H$  composed of states and actions, defined as $\sigma = \{s_k, a_k, \ldots, s_{k+H}, a_{k+H}\}$. 
The goal is to align human preference $y$ between pairs of trajectory segments $\sigma^0$ and $\sigma^1$, where $y$ denotes a distribution indicating human preference, captured as $y \in \{1, 0, 0.5\}$. 
The preference label $y=1$ indicates that $\sigma^0$ is preferred to $\sigma^1$, namely, $\sigma^0 \succ \sigma^1$, $y=0$ indicates $\sigma^1 \succ \sigma^0$, and $y=0.5$ indicates equal preference for both. 
The preference datasets are stored as triples, denoted as $\mathcal{D}_{pref}$: $(\sigma^0, \sigma^1, y)$.

The Bradley-Terry model \cite{bt95} is frequently employed to couple preferences with rewards. The preference predictor is defined as follows: 
\begin{equation}
    P_{\psi}[\sigma^1 \succ \sigma^0] = \frac{\exp\left(\sum_t \hat{r}_\psi(s_t^1, a_t^1)\right)}{\sum_{i \in \{0,1\}} \exp\left(\sum_t \hat{r}_\psi(s_t^i, a_t^i)\right)}
\end{equation}
where $\hat{r}_\psi$ is the reward model to be trained, and $\psi$ is its parameters. 
Subsequently, the reward function is optimized using the cross-entropy loss, incorporating the human ground-truth label $y$ and the preference predictor $P_{\psi}$:
\begin{equation}
\begin{aligned}
    \mathcal{L}_{\text{CE}} = -\mathbb{E}_{(\sigma^0,\sigma^1,y) \sim \mathcal{D}_{pref}}  & \Big\{ (1-y) \log P_{\psi}[\sigma^0 \succ \sigma^1] \\
    + & y \log P_{\psi}[\sigma^1 \succ \sigma^0] \Big\}
\label{eqt:pref_ce}
\end{aligned}
\end{equation}

In the offline PbRL, we assume there exists a small dataset $\mathcal{D}_{pref}$ with preference labels along with a much larger unlabeled dataset $\mathcal{D}$ without rewards or preference labels. 
We label the offline dataset $\mathcal{D}$ with estimated step-wise rewards $\hat{r}$. Then we can obtain the trajectory dataset  by calculating trajectory-wise RTG $\hat{R}$ for individual trajectories in  $\mathcal{D}$. The re-labeled dataset is used for downstream offline RL. 
The notation “in-dataset trajectory” indicates that the trajectory is in the offline dataset $\mathcal{D}$.

\subsection{Return-Based Conditional Sequence Modeling }
RL is formulated as a Markov Decision Process (MDP) \cite{rl18}.
A MDP is characterized by the tuple $M=\langle S, A, P, r,\gamma \rangle$, where $S$ is the state space, $A$ is the action space, $P: S \times A \times S \rightarrow \mathbb{R}$ is the transition probability distribution, $r: S \rightarrow \mathbb{R}$ is the reward function, and $\gamma \in (0, 1)$ is the discount factor.
The objective of RL is to determine an optimal policy $\pi^{}$ that maximizes the expected cumulative reward: $
\pi=\arg \max_\pi \mathbb{E}_{s_0, a_0, \ldots}\left[\sum_{t=0}^{\infty} \gamma^t r\left(s_t\right)\right]$.

{Return-based CSM addresses sequential decision problems through an autoregressive generative model, thus avoiding value estimation \cite{rcsl22}.} 
A prominent example is DT \cite{dt21}, which considers a sequence of trajectories of length $H$: $\tau_t = ({R}_{t-H+1}, s_{t-H+1}, a_{t-H+1}, \cdots, {R}_{t}, s_{t}, a_{t})$, where ${R}_{t}$ denotes return-to-go (RTG): $R_t = \sum_{t'=t}^T r_{t'}$. 
During the training phase, the supervised loss between the target policy and behavior policy is computed using mean squared error (MSE) loss.
During the inference phase, DT rolls out the estimated action $\hat{a}_t$ with the provided RTG  $\hat{R}_{t}$.
Although CSM harnesses the power of large models for supervised learning,
its limited trajectory stitching capability
constrains the potential for improvement from suboptimal data.

\section{Methods}
\begin{figure*}[h]
 \centering
 \includegraphics[width=1\linewidth]{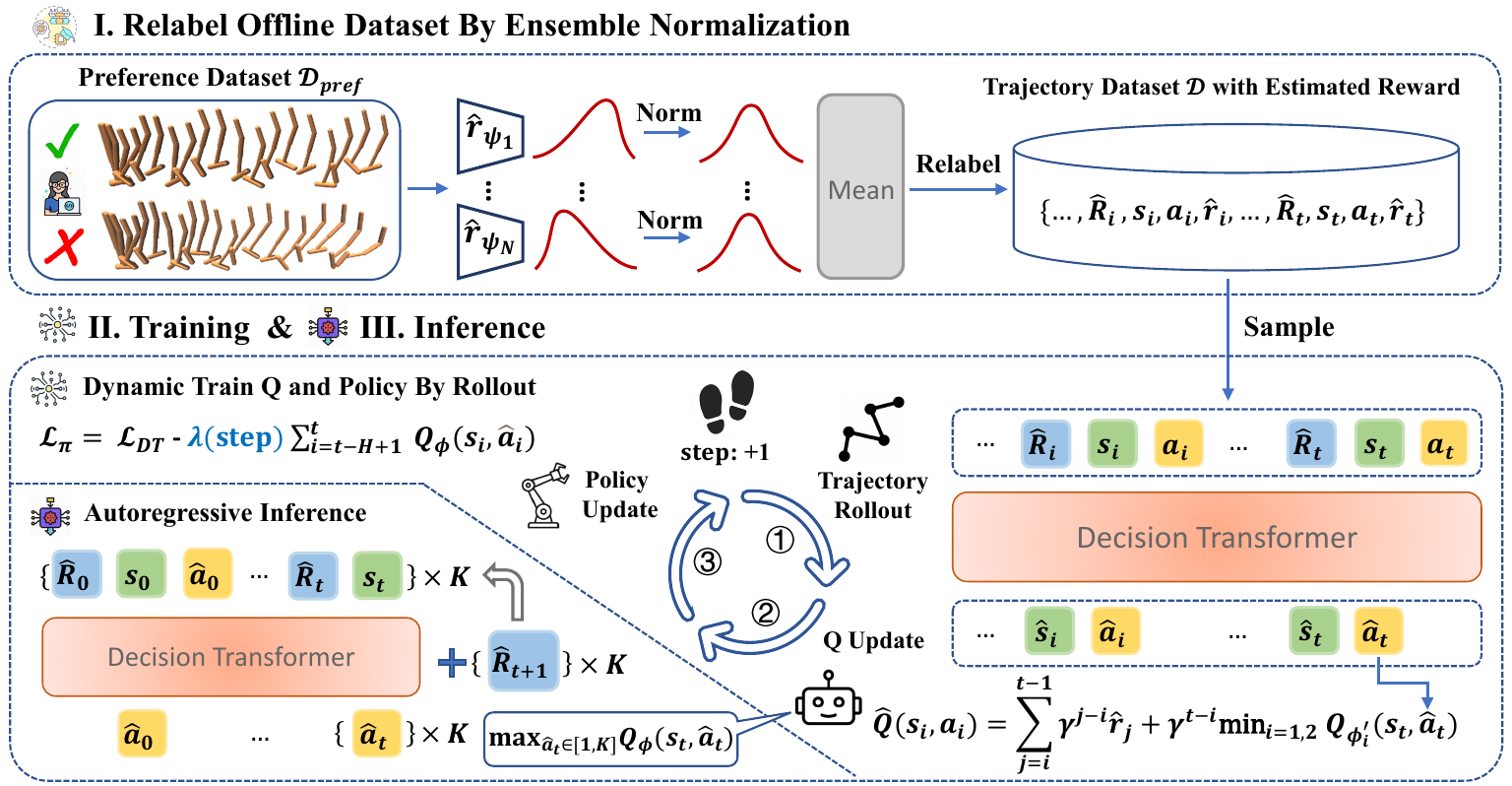}
 \caption{
 The overall framework DTR. 
 Phase I: Utilize preference data $\mathcal{D}_{pref}$
  to learn the reward model, then label the offline dataset $\mathcal{D}$ by ensemble normalization. 
  Phase II: Train the $Q$ function and policy network, comprising three components: trajectory rollout, $Q$ update, and policy update. 
  Phase III: Autoregressively infer actions based on the target RTG and select the action with the highest $Q$ value.
}
 \label{fig:main}
\end{figure*}
In this section, we first introduce the problem of reward bias in offline PbRL using a toy example and analyze the TDL and CSM methods in such scenarios to further emphasize our motivation.
Then, we provide a detailed pipeline and the overall framework for the DTR method.
Finally, we prove that extracting policies with in-dataset trajectory return regularization leads to {provable suboptimal bounds.}

\subsection{Rethinking Stitching in PbRL: A Toy Example}

We use the deterministic MDP depicted in Figure \ref{fig:mdp} as an example.
In this MDP, the agent starts from $s_1$ or $s_2$, transitions through $s_3$ or $s_4$, and finally reaches $s_5$ or $s_6$. 
The GT reward for each state transition is denoted as $r$. 
We estimate step-wise rewards $\hat{r}$ from  {trajectory-level} pairwise preferences. 
Typically, the preference dataset $\mathcal{D}_{pref}$ is gathered through a limited number of queries, making it difficult to cover all possible trajectories in the offline dataset $\mathcal{D}$. 
Suppose $\mathcal{D}_{pref}$ includes pairwise comparison labels for four trajectories and their intermediate segments:
$$
\mathcal{D}_{pref} = \left(
\begin{array}{l}
{(s_1, s_3, s_5)_\textit{red}},  \ \ \  {(s_1, s_4, s_6)_\textit{yellow}}, \  y = 1 \\
{(s_2, s_4, s_5)_\textit{green}}, {(s_1, s_4, s_6)_\textit{yellow}},  \ y = 1 \\
{(s_2, s_4, s_5)_\textit{green}}, {(s_1, s_3, s_5)_\textit{red}}, \ \ \ \ \  y = 1 \\
{(s_1, s_3, s_5)_\textit{red}}, \ \ \  {(s_2, s_4, s_6)_\textit{purple}}, \  y = 1
\end{array}
\right)
$$

\noindent Here we use $(\cdot)_{color}$ to denote the  trajectory and its color depicted in Figure \ref{fig:mdp}.
{Suppose the trajectories in $\mathcal{D}$ are identical to  those in $\mathcal{D}_{pref}$.}
There are three possible trajectories beginning from $s_1$: $(s_1, s_3,s_5)$, $(s_1,s_4,s_6)$ and $(s_1,s_4, s_5)$.  The first two are intrinsic in $\mathcal{D}$ (in-dataset), while the last one is formed by stitching together two trajectory fragments (out-dataset). 
The trajectory with the highest GT return is  $(s_1, s_3,s_5)$. 
However, due to the reward bias, we assume $\hat{r}_{45}=3$ is higher than the ground-truth ${r}_{45}=2$.
Consequently, the estimated return of the stitched trajectory $(s_1,s_4, s_5)$ is higher than that of $(s_1, s_3,s_5)$.
Such failures of estimated rewards are common when using a small-scale $\mathcal{D}_{pref}$ to {label a large-scale $\mathcal{D}$.}

We now scrutinize the performance of CSM and TDL methods under the reward bias. 
Starting from $s_1$, TDL chooses the stitched trajectory $(s_1,s_4, s_5)$ that maximizes the expected cumulative reward. 
In contrast, CSM identifies the in-dataset trajectory that maximizes the RTG for $s_1$, considering only trajectories within $\mathcal{D}$ that contain $s_1$, specifically $(s_1, s_3,s_5)$ and $(s_1, s_4,s_6)$.
As a result, CSM opts for $s_3$ over $s_4$, leading to $s_5$. 
To put it simply, CSM prefers conservative choices to learn behaviors under limited preference queries, {seeking optimal actions that align closely with the behavior policy conditioned on RTG. }
Conversely, TDL may fail due to optimistic trajectory stitching under reward bias.
Therefore, we hope to balance conservatism and exploration by integrating CSM and TDL dynamically for offline PbRL.

\subsection{In-dataset Regularization: Training and Inference}
In the offline policy training phase, we aim to utilize the DT and TD3 to achieve a balance between maintaining fidelity to the behavior policy with high in-dataset trajectory returns and selecting optimal actions with high reward labels.
Specifically, we sample a mini-batch trajectories from $\mathcal{D}$, and use DT as the policy network to rollout the estimated actions $\{\hat{a}_{i}\}_{i=t-H+1}^{t}$.
Due to the powerful capability of autoregressive generative model, we also rollout estimated states $\{\hat{s}_{i}\}_{i=t-H+1}^{t}$ to strengthen policy representation \cite{liu2024enhancing}.
Considering the trajectory $\tau_t=\{\cdots,\hat{R}_{i},{s}_{i},{a}_{i},\cdots,\hat{R}_{t},{s}_{t},{a}_{t}\}$,
the self-supervised loss of DT can be expressed as:
\begin{equation}
\mathcal{L}_{DT} = \mathbb{E}_{\tau_t \sim \mathcal{D}}  \sum_{i=t-H+1}^{t} \underbrace{\| \hat{s}_i - s_i \|^2}_{\text{Auxiliary Loss}} + \underbrace{\| \hat{a}_i - a_i \|^2}_{\text{Goal-BC Loss}}
\end{equation}

Additionally, we train the $Q$ function to select optimal actions with high reward labels. 
A natural idea is to use the estimated trajectory sequence for n-step TD bootstrapping.
Following \cite{td3_18} and \cite{qt24}, the $Q$ target and $Q$ loss are defined as follows:
\begin{equation}
\hat{Q}(s_i, a_i) = \sum_{j=i}^{t-1} \gamma^{j-i} \hat{r}_j + \gamma^{t-i} \min_{i=1,2} Q_{\phi_i}(s_t, \hat{a}_t)
\end{equation}
\begin{equation}
\mathcal{L}_Q = \mathbb{E}_{\tau_t \sim \mathcal{D}}  \sum_{i=t-H+1}^{t} \left\| Q_{\phi}(s_i, a_i) - \hat{Q}(s_i, a_i) \right\|^2
\label{eqt:n_step_Q}
\end{equation}

To prevent failures from incorrect trajectory stitching, as highlighted in the toy example, we dynamically integrate the policy gradient to train the policy network. 
Meanwhile, in Theorem 1, we will show that extracting policies from in-dataset trajectories leads to provable suboptimal bounds.
In the early training stage, the policy loss $\mathcal{L}_{\pi}$ primarily consists of supervised loss $\mathcal{L}_{DT}$ to ensure that the policy learns the trajectories of in-dataset optimal return  and the corresponding $Q$ function. 
Subsequently, we gradually increase the proportion of policy gradient to facilitate trajectory stitching near the optimal range within the distribution. 
In summary, we minimize the following policy loss:
\begin{equation}
 \mathcal{L}_\pi = \mathcal{L}_{DT} - \lambda(\text{step})  \ \mathbb{E}_{\tau_t \sim \mathcal{D} } \sum_{i=t-H+1}^{t} 
 % \frac{Q_{\phi}(s_i, \hat{a}_i)}{\mathbb{E}_{(s,a)\sim \tau_t}|Q_\phi(s,a)| }
 {Q_{\phi}(s_i, \hat{a}_i)}
\label{eqt:loss}
\end{equation}
where $\lambda(\text{step})$ considers the normalized $Q$ value and increases linearly with the training steps:
\begin{equation}
  \lambda(\text{step})=\frac{\eta}{\mathbb{E}_{(s,a) \sim \tau_t}|Q_\phi(s,a)|}, \  \eta = \frac{\text{step} \times \eta_{max}}{\text{max\_step}}
\label{eqt:para}
\end{equation}
Different  from  TD3BC \cite{TD3BC21} with the constant $\eta$, our approach uses progressively increasing coefficients to maintain training stability. This follows a straightforward principle: “Learn to walk before you can run.”

After the training, we get a trained policy $\pi_{\theta}$ and $Q$ network $Q_\phi$.
During the inference stage, we autoregressively rollout the action sequence almost following the setting of DT \cite{dt21}.
The difference is that we cannot update the RTG with the rewards provided by the online environment.
An alternative way is to replace the environment reward with a learned reward model:
\begin{equation}
    \hat{R}_{t+1} = \hat{R}_{t} - \hat{r}_{t}
    \label{eqt:r_update_origin}
\end{equation}
However, the reward model has limited generalization for OOD state-action pairs, and the additional model leads to greater consumption of computing resources.
Therefore, we propose a simple calculation: subtract a fixed value from the next timestep's RTG $\hat{R}_{t+1}$ until it is reduced to $0$ at the end of the episode:
\begin{equation}
    \hat{R}_{t+1} = \hat{R}_{t} - \hat{R}_{0} / \text{max\_timestep}
    \label{eqt:r_update}
\end{equation}
The reason for this simplification is that we normalize the rewards so that candidate states with high values tend to have similar rewards at each step. In implementation, we calculate $K$ initial RTGs as targets:
$\{\hat{R}_{0}^i\}_{i=1}^k=\{0.5, 0.75, 1, 1.5, 2\} \times \text{Return}_{\max}$, here we have $K=5$, and $\text{Return}_{\max}$ is the maximum return in $\mathcal{D}$.
Through the parallel reasoning of Transformer, $K$ actions are derived, and the one with the highest $Q$ value is executed.

\subsection{Relabel Offline Data by Ensemble Normalization}
During the annotation reward phase, we introduce a simple but effective normalization method to highlight reward differentiation, and further improve the performance of downstream policies. 
Specifically, we train $N$-ensemble MLP reward models $\{\hat{r}_{\psi_i}\}_{i=1}^{N}$ from the offline preference dataset $\mathcal{D}_{pref}$ with the loss function defined in Equation \ref{eqt:pref_ce}, then annotate the offline data $\mathcal{D}$ with predicted rewards.

We observe that the trained reward model labels different state-action pairs with only minor differences {(as detailed in the Experiments Section and Figure \ref{fig:abl_three1})},
and these indistinct reward signals may lead to exploration difficulties and low sample efficiency \cite{reward_exploration20}. 
While direct reward normalization can amplify these differences, it also increases uncertainty, resulting in inaccurate value estimation and high variance performance \cite{hpl24}. 

To balance these two aspects, we propose ensemble normalization, which first normalizes the estimates of each ensemble and then averages them:
\begin{equation}
    \hat{r}_{\psi} = \text{Mean}(\text{Norm}(\hat{r}_{\psi_1}),\cdots,\text{Norm}(\hat{r}_{\psi_N}))
\label{eqt:reward_norm}
\end{equation}
Notably, ensemble normalization is a plug-and-play module that can enhance reward estimation without any modifications to reward training. 

\subsection{Theoretical Analysis}

Suppose the preference dataset $\mathcal{D}_{pref}$ with $N_p$ trajectories of length $H_p$ and the {reward-free offline dataset $\mathcal{D}$} with $N_o$ trajectories of length $H_o$.
Then, consider the following general offline PbRL algorithm: 

\noindent \textbf{1. Construct the Reward Confidence Set: }

First, estimate the parameters ${\widehat{\psi}} \in \mathbb{R}^d$ of reward model  via MLE with Equation \ref{eqt:pref_ce}. Then, construct a confidence set $\Psi(\mathcal{\zeta})$ by selecting reward models that nearly maximize the log-likelihood of $\mathcal{D}_{pref}$ to a slackness parameter $\zeta$. 

\noindent  \textbf{2. Bi-Level Optimization of Reward and Policy:}

Identify the policy $ \widehat{\pi}$ that maximizes the estimated policy value $\widehat{V}$ under the least favorable reward model $\widetilde{\psi}$ with both $\mathcal{D}_{pref}$ and $\mathcal{D}$:
$$\widetilde{\psi} = \arg \min_{\psi \in \Psi(\mathcal{\zeta})} {\widehat{V}_\psi - \mathbb{E}_{\tau \sim \mu_{ref}} [r_\psi(\tau)] }, \ \ \widehat{\pi} = \arg \max_\pi \widehat{V}_{\widetilde{\psi}}$$
And we have the following theorem:    
\begin{theorem}[Informal]
Suppose that: 
(1) the dataset $\mathcal{D}_{pref}$ and $\mathcal{D}$ have positive coverage coefficients $C_p^\dagger$ and $C_o^\dagger$;
(2) the underlying MDP is a linear MDP;
(3) the GT reward $\psi^\star \in \mathcal{G}_\psi$; 
(4) $0 \leq r_\psi(\tau) \leq r_{\max} $, and $\|\psi\|_2^2 \leq d$ for all $\psi \in \mathcal{G}_\psi$ and $\tau \in \mathcal{T}$.
and 
(5) $\widehat{\pi}$ is any mesurable function of the data $\mathcal{D}_{pref}$.
Then with probability $1-2 \delta$, the performance bound of the policy $\widehat{\pi}$ satisfies for all $s \in \mathcal{S}$,

\begin{equation}
\begin{aligned}
\text{SubOpt}(\widehat{\pi}; s) &\le
\sqrt{\frac{c C_\psi^2(\mathcal{G}_\psi, \pi^{\star}, \mu_{ref}) \kappa^2 \log(\mathcal{N}_{\mathcal{G}_\psi} / N_p\delta)}{N_p}} \\
&+
\frac{2c r_{\max}}{(1 - \gamma)^2} \sqrt{\frac{d^3 \xi_\delta}{N_p H_p C_p^\dagger + N_o H_o C_o^\dagger}}    
\label{eq:subopt}
\end{aligned}
\end{equation}
\label{theorem1}
\end{theorem}
where $\xi_\delta=\log \left({4d(N_p H_p+N_o H_o)}/{(1-\gamma) \delta} \right)$, and other variables involved are not related to with $N_p$ or $N_o$. For detailed definitions, see Appendix A.

\begin{remark}
The theorem extends offline RL theory specifically to offline PbRL, leading to a theoretical upper bound for offline PbRL.
In order to guarantee this bound, the learned policy $\widehat \pi$ is any measurable function of  $ \mathcal{D}_{pref}$ should be satisfied. 
In other words, if $\widehat \pi \in \mathcal{D}_{pref}$, the assumption naturally holds.
A relaxed condition is $\widehat \pi \in \mathcal{D}$, since trajectories in $ \mathcal{D}_{pref}$ are often sampled from $\mathcal{D}$.
This theoretical analysis guides us to make more use of in-dataset trajectories for policy optimization to ensure marginal performance improvements. 
Accordingly, our DTR method utilizes CSM for optimizing in-dataset trajectories to establish reliable performance bound, while employing TDL to enhance the utilization of  out-dataset trajectories.
We elaborate on this finding with experiments in Appendix E7.
\end{remark}

\section{Experiments}
\begin{table*}[h]
\centering
\begin{adjustbox}{width=\textwidth}
\begin{tabular}{lccccc|cc|c c}
\toprule
 \textbf{Dataset} & \textbf{Pb-IQL} & \textbf{Pb-TD3BC} & \textbf{DPPO} & \textbf{OPRL} & \textbf{FTB} & \textbf{Pb-DT*} & \textbf{Pb-QT*} & \textbf{DTR (Ours)*} & \textbf{DTR (Best)*} \\ \midrule
    Walker2D-m & 78.4 & 26.3 & 28.4 & \underline{80.8} & 79.7 & 71.4 ± {\scriptsize 5.1} & 80.1 ± {\scriptsize 1.4}& \textbf{86.6} ± {\scriptsize 2.8}& 88.3 ± {\scriptsize 2.7} \\ 
    Walker2D-m-r & 67.3 & 47.2 & 50.9 & 63.2 & \underline{79.9}& 51.1 ± {\scriptsize 10.5} & {79.3} ± {\scriptsize 1.5}& \textbf{80.8} ± {\scriptsize 2.8}& 84.4 ± {\scriptsize 2.1} \\ 
    Walker2D-m-e & {109.4} & 74.5 & {108.6} & \underline{109.6} & {109.1} & {108.0} ± {\scriptsize 0.2} & {109.5} ± {\scriptsize 0.6}& \textbf{109.7} ± {\scriptsize 0.3} & 111.1 ± {\scriptsize 0.3} \\ 
    Hopper-m & 50.8 & 48.0 & 44.0 & 59.8 & 61.9 & 49.8 ± {\scriptsize 4.8} &\underline{81.3} ± {\scriptsize 8.8}& \textbf{90.7} ± {\scriptsize 0.6} & 94.5 ± {\scriptsize 0.4} \\ 
    Hopper-m-r & 87.1 & 25.8 & 73.2 & 72.8 & \underline{90.8} & 67.0 ± {\scriptsize 8.2} & 84.5 ± {\scriptsize 12.8}& \textbf{92.5} ± {\scriptsize 0.9} & 96.0 ± {\scriptsize 1.6} \\ 
    Hopper-m-e & 94.3 & 97.4 & {107.2} & 81.4 & \underline{110.0} & \textbf{111.3} ± {\scriptsize 0.1} & {109.4} ± {\scriptsize 1.8}& {109.5} ± {\scriptsize 2.4} & 112.3 ± {\scriptsize 0.3} \\ 
    Halfcheetah-m & 43.3 & 34.8 & 38.5 & \textbf{47.5} & 35.1 & 42.5 ± {\scriptsize 0.8} & 42.7 ± {\scriptsize 0.3}& \underline{43.6} ± {\scriptsize 0.3}& 44.2 ± {\scriptsize 0.3}\\ 
    Halfcheetah-m-r & 38.0 & 38.9 & \underline{40.8} & \textbf{42.3} & 39.0 & 37.6 ± {\scriptsize 0.6} & 39.9 ± {\scriptsize 0.7}& {40.6} ± {\scriptsize 0.2}& 41.6 ± {\scriptsize 0.2}\\ 
    Halfcheetah-m-e & {91.0} & 73.8 & \textbf{92.6} & 87.7 & {91.3} & 68.7 ± {\scriptsize 9.1} & 78.2 ± {\scriptsize 11.6}& \underline{91.9} ± {\scriptsize 0.4}& 93.5 ± {\scriptsize 0.3}\\ \midrule
    \textbf{Average} & 73.29 & 51.86 & 64.91 & 71.68 & 77.42 & 67.49 & \underline{78.32}& \textbf{82.88}& 85.10\\ \bottomrule
\end{tabular}
\end{adjustbox}
\caption{The performance comparison of DTR and different baselines on Gym-MuJoCo locomotion. 
In the first column, -m, -m-r, and -m-e are abbreviations for the medium, medium-replay, and medium-expert datasets, respectively. 
The results of Pb-IQL and Pb-TD3BC are from Uni-RLHF benchmark \cite{uni-rlhf}, the results of OPRL and FTB are from the experimental section of FTB \cite{ftb24}, the -m-r and -m-e results of DPPO are from \cite{dppo23}, and the -m results are reproduced by ourselves. 
For the remaining methods with $\textbf{*}$, we record the average normalized score of the 10 rollouts of the last checkpoint and run the experiment under 5 random seeds, and finally record the mean score and ± denotes the standard deviation. The
\textbf{bold font} indicates the algorithm with the best performance, and the \underline{underline} indicates the second one.
}
\label{tab:mujoco}
\end{table*}

\begin{table}[h]
\centering
\begin{adjustbox}{width=\columnwidth}
\begin{tabular}{lcccc}
\toprule
\textbf{Dataset} & \textbf{Pb-IQL} & \textbf{Pb-DT} & \textbf{Pb-QT} & \textbf{DTR (Ours)} \\ \midrule
    Pen-h & 99.8 ± {\scriptsize 8.3} & \textbf{115.3} ± {\scriptsize 2.0} & 111.2 ± {\scriptsize 3.9} & \underline{114.0} ± {\scriptsize 9.6} \\ 
    Pen-c & \underline{93.7} ± {\scriptsize 14.1} & \textbf{104.6} ± {\scriptsize 13.4} & 69.1 ± {\scriptsize 12.0} & 86.6 ± {\scriptsize 6.4} \\ 
    Door-h & 9.7 ± {\scriptsize 1.4} & 14.2 ± {\scriptsize 1.4} & \underline{26.0} ± {\scriptsize 5.0} & \textbf{33.3} ± {\scriptsize 8.6} \\ 
    Door-c & 2.2 ± {\scriptsize 0.6} & 7.7 ± {\scriptsize 0.9} & \underline{10.7} ± {\scriptsize 1.9} & \textbf{12.6} ± {\scriptsize 0.6} \\ 
    Hammer-h & 11.8 ± {\scriptsize 2.0} & 2.0 ± {\scriptsize 0.3} & \underline{15.0} ± {\scriptsize 5.7} & \textbf{21.5} ± {\scriptsize 6.9} \\ 
    Hammer-c & 11.4 ± {\scriptsize 2.7} & 4.0 ± {\scriptsize 2.1} & \underline{14.6} ± {\scriptsize 1.5} & \textbf{26.7} ± {\scriptsize 10.3} \\ \midrule
    \textbf{Average} & 38.10 & \underline{41.29} & 41.10 & \textbf{49.11} \\ \bottomrule
\end{tabular}
\end{adjustbox}
\caption{The performance comparison on Adroit manipulation platform. In the first column, -h and -c are abbreviations for the human and clone datasets. 
}
\label{tab:adroit}
\end{table}

In this section, we evaluate DTR and other baselines on various benchmarks. 
We aim to address the following questions:
(1) Can DTR mitigate the risk of reward bias and lead to enhanced performance in offline PbRL?
(2) What specific roles do the proposed modules play in the proposed DTR? 
(3) Can DTR still perform well with small amounts of preference feedback?

\subsubsection{Setup.}
We select three tasks from the Gym-MuJoCo locomotion suite \cite{openaigym}, and
three tasks from the Adroit manipulation platform \cite{Kumar2016thesis}.
We utilize the Uni-RLHF dataset \cite{uni-rlhf} as preference dataset $\mathcal{D}_{pref}$, which provides pairwise preference labels from crowdsourced human annotators. 
The offline dataset $\mathcal{D}$ is sourced from the D4RL benchmark \cite{d4rl20}.

\subsubsection{Baselines.} 
(1) Pb \cite{pt23} including PT-IQL and Pb-TD3BC, which perform IQL or TD3BC with preference-based trained reward functions, respectively; 
(2) DPPO \cite{dppo23}, which designs a novel policy scoring metric and optimize policies without reward modeling;
(3) OPRL \cite{oprl23}, which performs IQL with ensemble-diversified reward functions;
(4) FTB \cite{ftb24}, which generates better trajectories with higher preferences based on diffusion model.
For a more intuitive comparison, we propose several variants:
(5) Pb-DT and Pb-QT, which slightly modify CSM models DT \cite{dt21} and QT \cite{qt24} for offline PbRL. The more implementation details are shown in Appendix D. 

It is worth emphasizing that we provide a structural comparison of reward model between MLP and Transformer for offline PbRL. 
We find that that MLP-based reward model has a more stable performance than the Transformer-based one. 
Please see Appendix E2 for the detailed analysis. 
Hence, our methods employ the MLP-based reward model. 

\begin{figure}[t]
\centering
\subfloat[Ablation study on $\eta$.]{
\includegraphics[width=.502\linewidth]{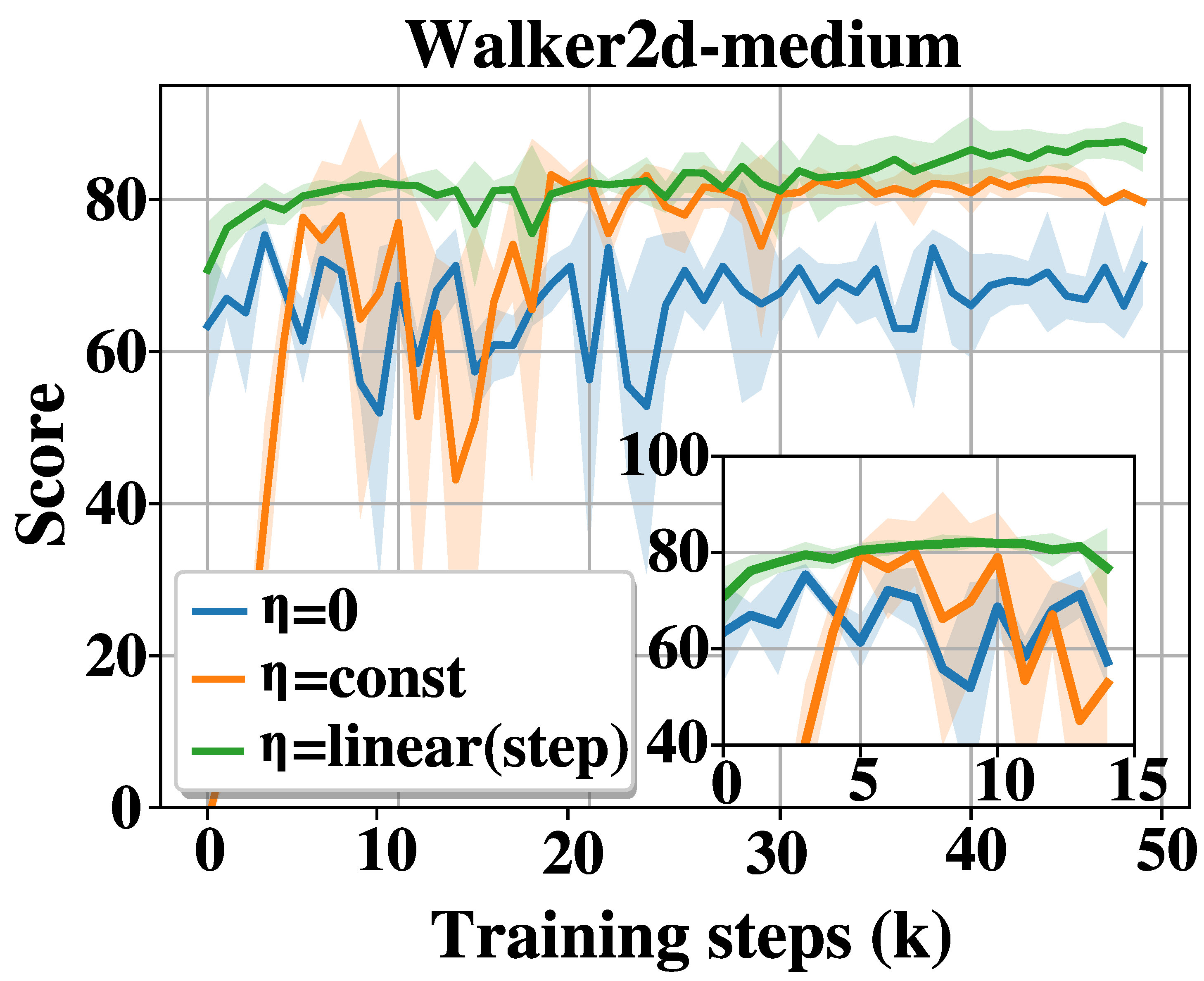}
\label{fig:abl_both1}
}
% \hfill
\subfloat[Ablation study on norm\_r.]{
\includegraphics[width=.472\linewidth]{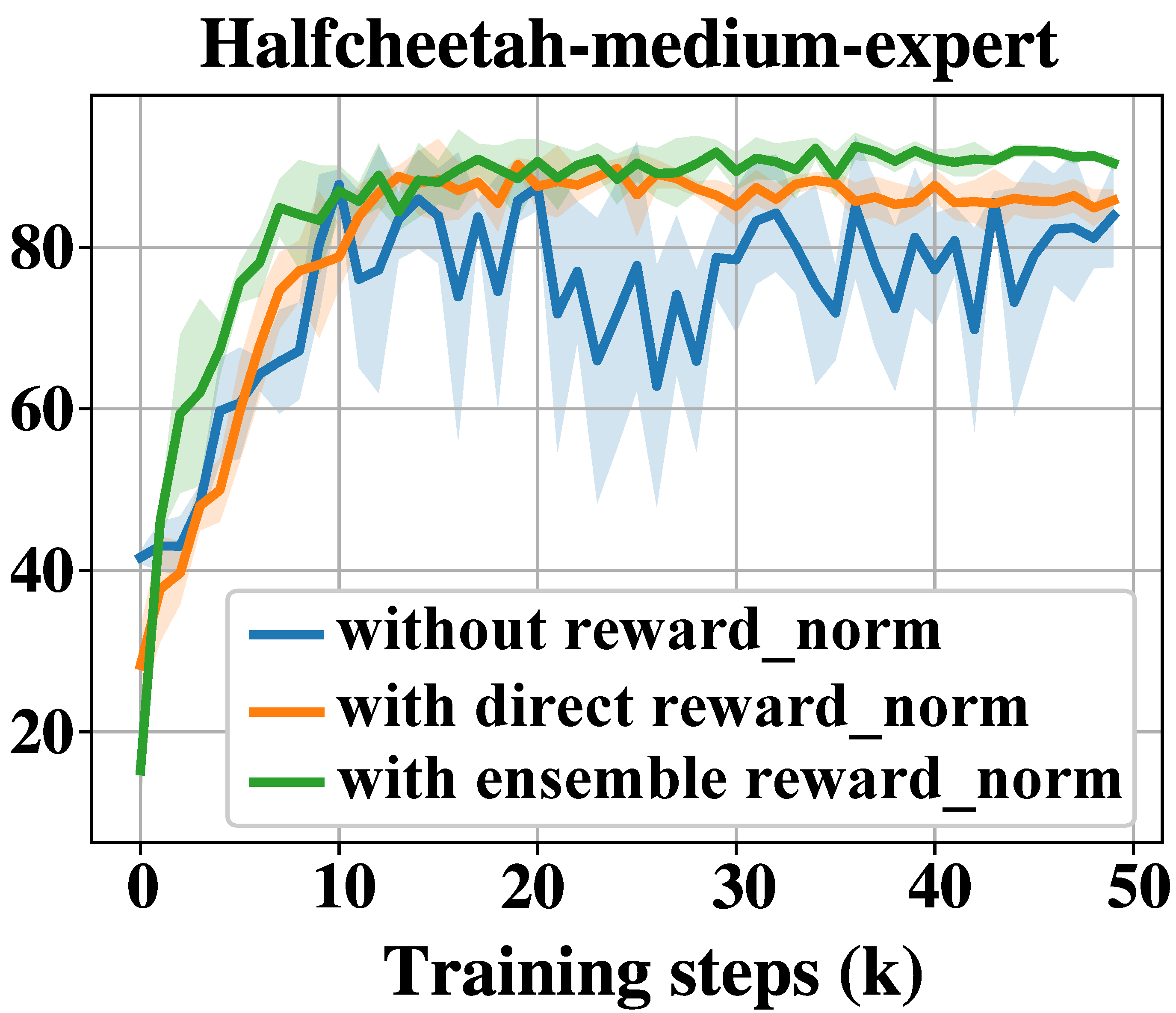}
\label{fig:abl_both2}
}
\caption{Training score curve for the ablation study.}
\label{fig:abl_both}
\end{figure}

\begin{figure*}[h]
\centering
\subfloat[Compare of Reward Normalization.]{
\includegraphics[width=.34\linewidth]{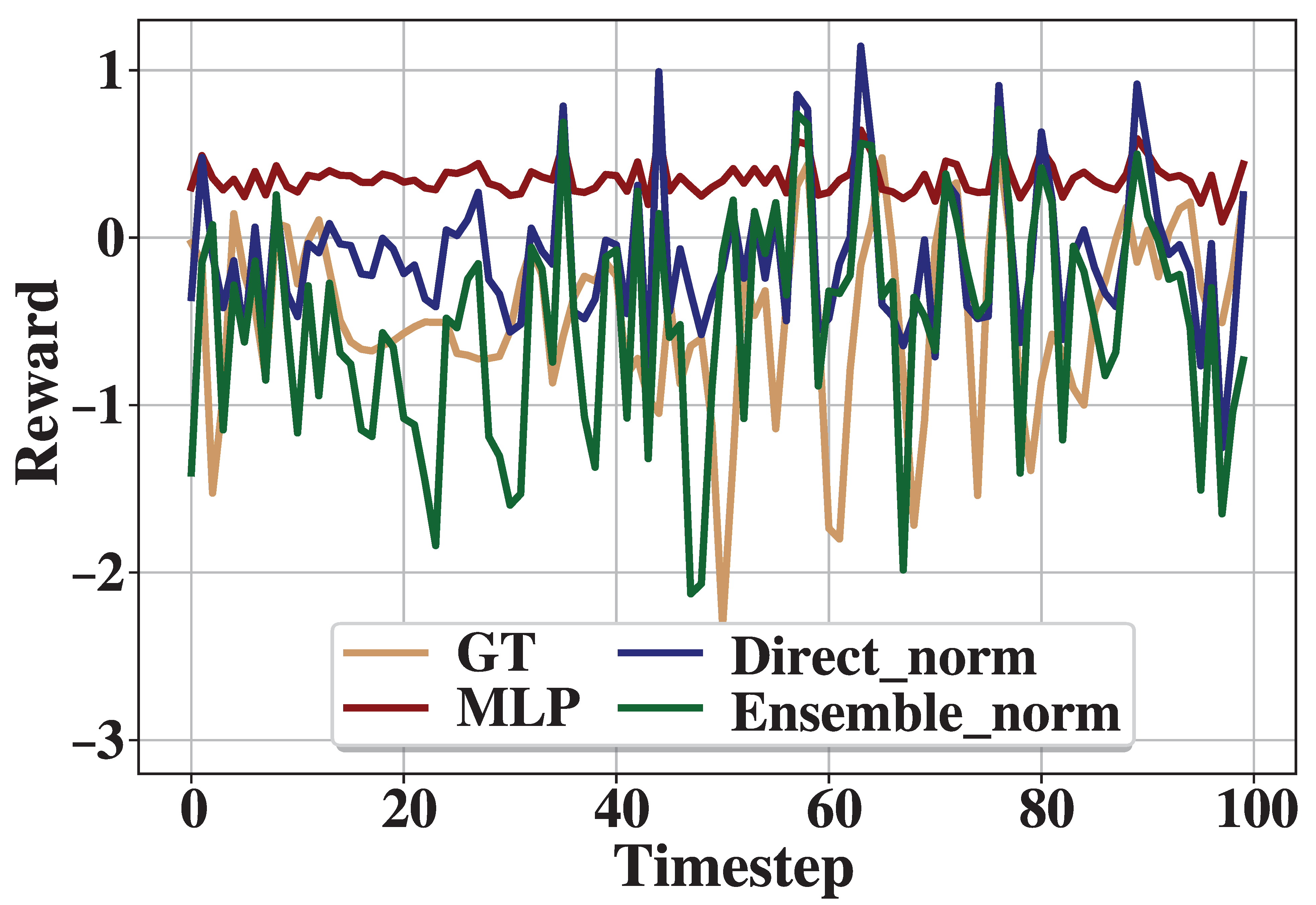}
\label{fig:abl_three1}
}
% \hfill
\subfloat[Ablation with Ground-Truth Reward.]{
\includegraphics[width=.3\linewidth]{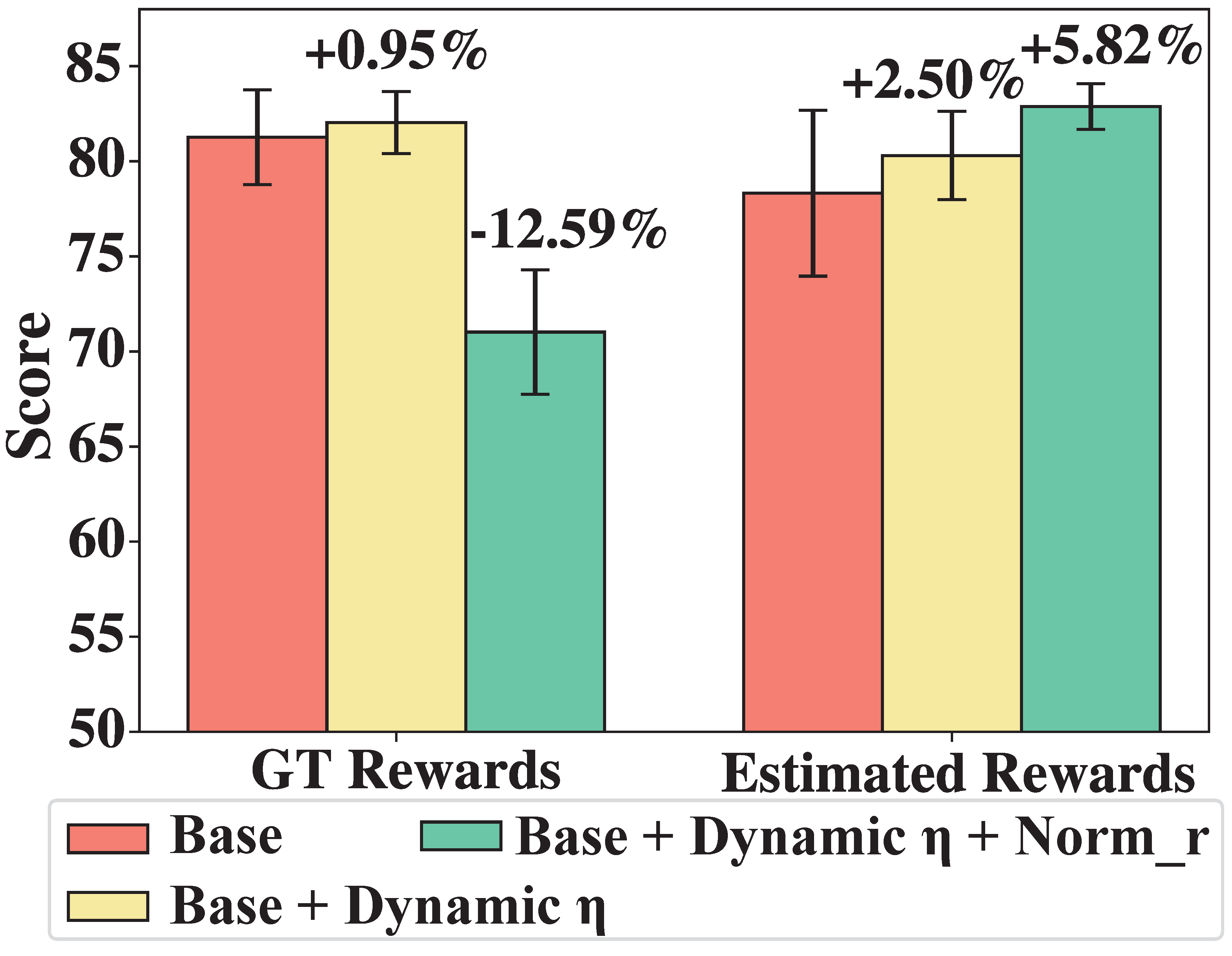}
\label{fig:abl_three2}
}
% \hfill
\subfloat[Scaling Trend for Preference Dataset.]{
\includegraphics[width=.325\linewidth]{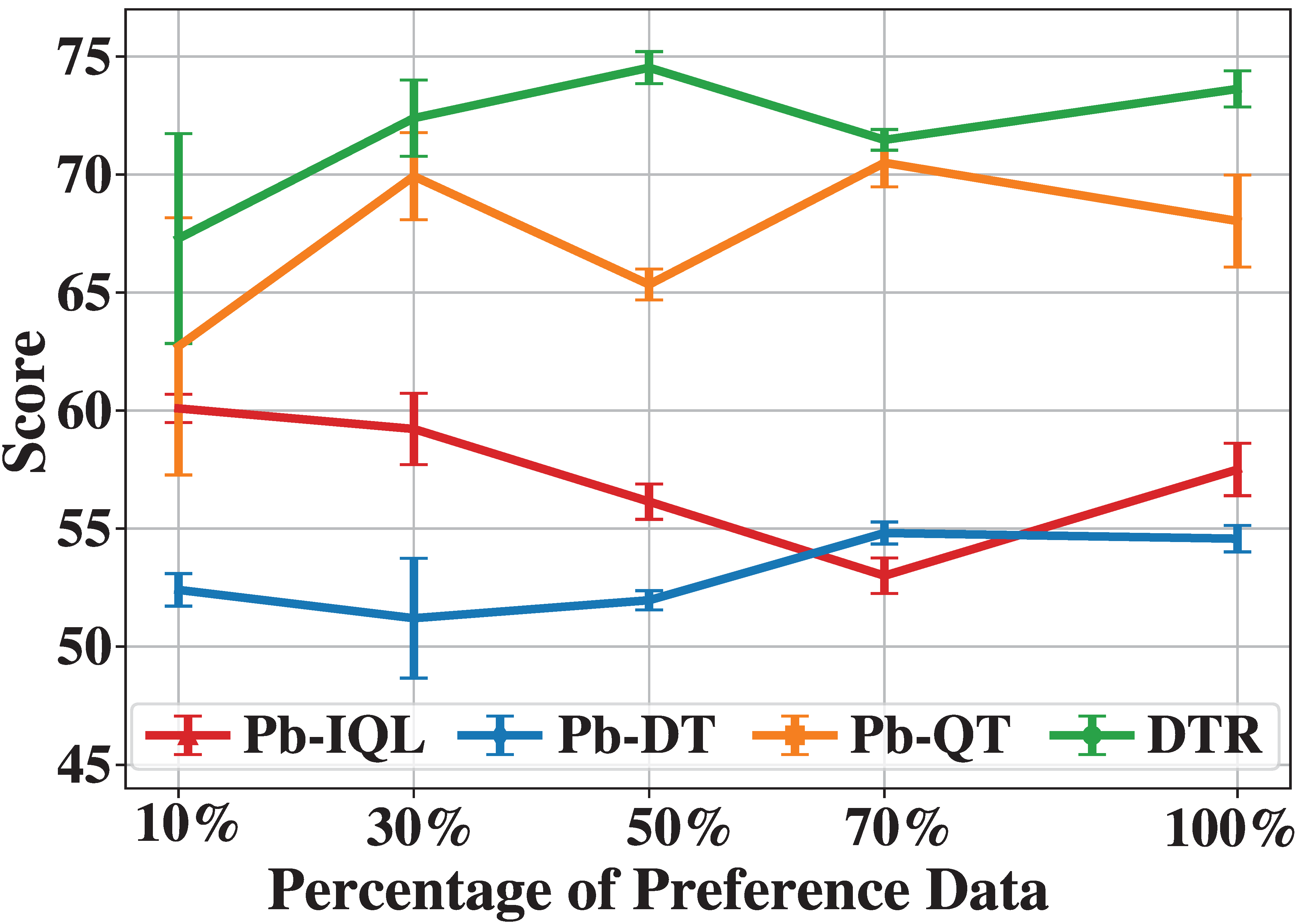}
\label{fig:abl_three3}
}
\caption{
    Visual comparison of ablation experiments. 
    (a) Changes in the reward values of trajectories labeled with different reward normalization methods in Halfcheetah-m-e; 
    (b) The impact of the proposed module on the performance under the GT rewards and estimated rewards datasets. We record the average score across all 9 tasks of MuJoCo and "Base" represents DTR without the dynamic coefficient and ensemble normalization; 
    (c) The performance of different methods changes with the size of the preference dataset. We record the average score across 3 tasks of MuJoCo-Medium.
}
\label{fig:abl_three}
\end{figure*}

\subsection{Q1: Can DTR Mitigate the Risk of Reward Bias and Lead to Enhanced Performance?}

\subsubsection{Gym-MuJoCo Locomotion Tasks.} 
In the MuJoCo environment, we comprehensively compare DTR with the above baselines.
The results are listed in Table \ref{tab:mujoco}. 
DTR outperforms all prior methods in most datasets. On average, DTR surpasses the best baseline FTB by a large margin with a minimum of 7.1\%  and even 11.0\% in our best performance.
Additionally, DTR exhibits significantly lower performance variance compared to Pb-DT and Pb-QT, which suffer from pronounced fluctuations. 
{The superior performance  reflects the  importance of integrating CSM and TDL to mitigate the potential risk of reward bias in offline PbRL.}

\subsubsection{Adroit Manipulation Platform.} 
We further evaluate DTR on the more challenging Adroit manipulation platform. 
We mainly compare DTR with Pb-IQL since it performs best on this benchmark in previous work \cite{uni-rlhf}. 
% We observe that the scores of all methods fluctuated during training, so we evaluate at regular intervals and recorded the maximum score among all evaluations. 
% DTR outperforms Pb-IQL on difficult tasks 
% (e.g., Hammer and Door), 
with an average score higher than IQL by 28.9\%.

\subsection{Q2: What Specific Roles Do the Proposed Modules Play in DTR?}

\subsubsection{The Role of Dynamic Coefficient.}
To further elucidate  the role of the proposed dynamic coefficients $\lambda(\text{step})$ in Equation \ref{eqt:loss} and \ref{eqt:para}. 
We study the performance difference compared to $\eta=0$ and $\eta=\text{const}$,
and present the training score curves in Figure \ref{fig:abl_both1}. 
The results of first 15k training steps highlight the superior training efficiency of dynamic coefficient, while the method with constant $\eta$ fluctuates and performs lower than the dynamic coefficient one. 

\subsubsection{Comparison of Reward Normalization.} 
To explore the effectiveness of the proposed ensemble normalization of rewards, 
we present the training score curves for different normalization methods in Figure \ref{fig:abl_both2}.
The curves demonstrate that the ensemble normalization method achieves superior convergence and reduced training fluctuations. 
Furthermore, we select the first 100 timesteps of the trajectory with the lowest GT return in Halfcheetah-m-e dataset and plot the changes in different estimated rewards in Figure \ref{fig:abl_three1}.
Compared to GT rewards, the original rewards (MLP) have little differences across states, which is detrimental to downstream policy learning.
Directly normalizing average rewards amplifies these differences but also exaggerates estimation errors, especially overestimation.
Our method normalizes the estimates of each ensemble member individually before averaging them, thereby enhancing reward differences while mitigating estimation errors.

\subsubsection{Overall Ablation Performance.}  Table \ref{tab:abl} records the average scores of the above two ablation experiments, and more ablation results are given in Appendix E5.

\subsubsection{Ablation with GT Reward.} 

To explore the ability of DTR in the offline RL domain, we apply the same dynamic coefficient and normalization modules as DTR on the offline dataset with GT rewards and {show the results in Figure \ref{fig:abl_three2}.} 
With GT rewards, the method incorporating the dynamic coefficient $\eta$ performs closely to the oracle method.
However, the performance significantly drops with reward normalization, likely due to inaccurate value estimation caused by the normalization of precise rewards.
DTR, which employs preference learning to estimate rewards, achieves performance comparable to the oracle.

\begin{table}[t]
\centering
\begin{adjustbox}{width=\columnwidth}
\begin{tabular}{lccc}
\toprule
\textbf{Gym-MuJoCo} & \textbf{DTR} & \textbf{-dynamic $\eta$} & \textbf{-reward\_norm} \\ \midrule
    \textbf{Average} & 82.88 ± {\scriptsize 1.20}  &  79.63 ± {\scriptsize 2.27} (-3.9\%) &    80.30 ± {\scriptsize 0.77} (-3.1\%) \\ \bottomrule
\end{tabular}
\end{adjustbox}
\caption{The Gym-MoJoCo average score of DTR without dynamic coefficient ($\eta=\text{const}$) and reward normalization.
}
\label{tab:abl}
\end{table}

\subsection{Q3: Can DTR Still Perform Well With Small Amounts of Preference Feedback?}

Human preference queries are often limited, so effective algorithms should align with human intentions using fewer queries. 
Consequently, we conduct experiments to evaluate the scalability of DTR and baselines with {varying preference dataset scales in the MuJoCo-medium environment.} 
We respectively train the reward model with portions of the queries from a dataset of 2000 preference pairs and then train downstream offline RL.  
The evaluation results are reported in Figure \ref{fig:abl_three3}.
It shows that DTR can also achieve good performance when reducing the amount of preference data.
indicating that finer reward estimation aids in enhancing the performance of the TDL module.

\section{Conclusion}
In this work, 
we propose DTR, which dynamically combines DT and TDL and utilizes the in-dataset trajectory returns and ensemble reward normalization to alleviate the risk of reward overestimation in offline PbRL.
% Evaluation and ablation experiments on public benchmarks demonstrate the excellent performance of DTR.
We show the potential of CSM in offline PbRL, which may motivate further research on generative methods in this field and extend them to online settings.
% One limitation is that we did not evaluate the inaccuracy of rewards and iteratively optimize the reward model in downstream policy.
In the future, combining the bi-level optimization of the reward function and the policy will be  an interesting topic.

\clearpage
\section{Acknowledgements} 
This work is supported by the National Key Research and Development Program of China under Grants 2022YFA1004000, the National Natural Science Foundation of China under Grants 62173325, and the CAS for Grand Challenges under Grants 104GJHZ2022013GC.

\bibliography{aaai25}

\clearpage
\onecolumn
\appendix

\begin{center}
    \LARGE \textbf{Appendix}
\end{center}

\section{A. Complete Theories and Proofs}
\subsection{A1. Problem Settings and Symbol Description} \

We focus on the problem of offline PbRL in the trajectory-based pairwise comparison setting, we are provided with an offline preference dataset $\mathcal{D}_{pref} = \left\{ \left( \tau^{n,0}, \tau^{n,1}, o^n \right) \right\}_{n=1}^{N_p}$, where $\tau^{n,0} = \left\{ s_h^{n,0}, a_h^{n,0} \right\}_{h=1}^{H_p}$ and $\tau^{n,1} = \left\{ s_h^{n,1}, a_h^{n,1} \right\}_{h=1}^{H_p}$ are i.i.d. sampled from the distributions $\mu_0$ and $\mu_1$, respectively, and $o^n \in \{0, 1\}$ indicates preference for $\tau^{n,1}$ over $\tau^{n,2}$.
% Define the number of transitions in preference dataset $N_p=N\times H$ and the coverage coefficient is $C_p^\dagger$.
In addition, we also have an unlabeled offline dataset $\mathcal{D} = \left\{ \left( \tau^{n} \right) \right\}_{n=1}^{N_o}$ with $\tau^{n} = \left\{ s_h^{n}, a_h^{n} \right\}_{h=1}^{H_o}$.

% the coverage coefficient $C_o^\dagger$. 

First, we clarify some concepts, which will be mentioned repeatedly: 

\begin{itemize}
\item $\psi \in \mathbb{R}^{d}$ is the parameter of the reward model, $V_{\psi}^{\pi}(s)$ represents the ground-truth value of $s$ with the policy $\pi$ and the reward model $r_\psi$, that is, $V_{\psi}^{\pi}(s) = \mathbb{E}_{\tau \sim \pi} [r_{\psi}]$;
\item $r_{\psi^\star}$ represents the optimal (GT) reward function, corresponding to the value function $V_{\psi^{\star}}$;
\item $r_{\widehat{\psi}}$ represents the reward function estimated by Maximum Likelihood Estimation (MLE), corresponding to  $V_{\widehat{\psi}}$;
\item $r_{\widetilde{\psi}}$ represents the reward function estimated by pessimistic  MLE, corresponding to $V_{\widetilde{\psi}}$;
\item $\widehat{V}$  represents the value estimation of downstream offline RL algorithm;
\end{itemize}

According to \cite{theory23}, when we consider the performance of the induced policy, MLE $r_{\widehat{\psi}}$ obviously fails, while the pessimistic MLE $r_{\widetilde{\psi}}$ gives a rate close to the optimal. 
Essentially, the pessimistic principle ignores actions that are less common in the observed dataset and is therefore conservative in outputting the policy.
Although, in practice we did not use the pessimistic MLE policy, in fact, most previous PbRL methods did not guarantee reward pessimism in theory. 
The reason is that pessimistic MLE requires considering both the reward and the policy during optimization, but with the neural network approximation, the confidence set of the reward model is difficult to construct.
However,  we still need pessimistic MLE as a guarantee in theory, so we propose the following general offline PbRL algorithm assumption:
\begin{assumption}[General offline PbRL Algorithm] \

First, we estimate the reward model $r_{\widehat{\psi}}$ by MLE. 
Then, we construct a confidence set for the ground-truth reward from the implicit preference feedback. We achieve this by selecting reward models that nearly maximize the log-likelihood of observed data up to a slackness parameter $\zeta$:

\begin{itemize}
    \item \textbf{MLE}:  $\widehat{\psi} = \arg\max_{\psi \in \mathcal{G}_\psi} \sum_{n=1}^N \log P_\psi (o = o_n \mid \tau_n^1, \tau_n^0)$

    \item \textbf{Confidence Set Construction}: $\Psi(\mathcal{\zeta}) = \left\{\psi \in \mathcal{G}_\psi: \sum_{n=1}^N \log P_\psi (o = o^n \mid \tau_n^0, \tau_n^1) \geq \sum_{n=1}^N \log P_{\widehat{\psi}} (o = o^n \mid \tau_n^0, \tau_n^1) - \zeta \right\}$
\end{itemize}
where $\mathcal{G}_\psi$ is a function class   such as linear functions or neural networks, to approximate the true reward; $P(o = 1 \mid \tau^0, \tau^1) = P(\tau^1 \text{ is preferred over } \tau^0 \mid \tau^0, \tau^1) = \Phi(r_{\psi^\star}(\tau^1) - r_{\psi^\star}(\tau^0))$ and $\Phi : \mathbb{R} \to [0,1]$ is a monotonically increasing link function.

Upon constructing the confidence set, we identify the policy that maximizes the policy value under the least favorable reward model with both $\mathcal{D}_{pref}$ and $\mathcal{D}$. Downstream offline RL is formulated as a two-level optimization problem, wherein the parameters of the reward model and the policy  are interdependent:

\begin{itemize}
    \item \textbf{Pessimistic  Reward Estimation}: $\widetilde{\psi} = \arg \min_{\psi \in \Psi(\mathcal{\zeta})} {\widehat{V}_\psi - \mathbb{E}_{\tau \sim \mu_{ref}} [r_\psi(\tau)] }$
    \item \textbf{Policy Estimation}: $\widehat{\pi} = \arg \max_\pi \widehat{V}_{\widetilde{\psi}}$
\end{itemize}
The reference policy $ \mu_{ref}$ is essential, as the approximated confidence set measures the uncertainty in reward differences between two trajectories $r(\tau^1)-r(\tau^0)$, but it cannot measure the uncertainty of the reward for a single trajectory.
\label{assumption1}
\end{assumption}

\begin{assumption}[Realizability \cite{provable24}]
    We have $\psi^\star \in \mathcal{G}_\psi$.
\label{assumption2}
\end{assumption}

\begin{assumption}[Boundedness \cite{hu2023provable}]
    We have $0 \leq r_\psi(\tau) \leq r_{\max}$ and $\|\psi\|_2^2 \leq d$ for all $\psi \in \mathcal{G}_\psi$ and $\tau \in \mathcal{T}$.
\label{assumption3}
\end{assumption}

\subsubsection{A2. Existing Lemmas} \

Next, we provide some existing lemmas. For detailed proofs, please refer to the original papers cited.

\begin{lemma}[Performance of MLE (In \cite{provable24}, Lemma 1)] \ 

Let $\zeta = c_{MLE} \log( {\mathcal{N}_{\mathcal{G}_\psi}}/{N_p\delta}) $ and $\mathcal{E}_a$ be the event that the
following inequality holds for any $\delta \in (0,1)$,
\begin{equation}
\sum_{n=1}^{N_p} \log \left( \frac{P_\psi(o^n \mid \tau_n^0, \tau_n^1)}{P_{\psi^\star}(o^n \mid \tau_n^0, \tau_n^1)} \right) \leq c_{MLE} \log \left( \frac{\mathcal{N}_{\mathcal{G}_\psi}}{N_p\delta} \right)
\end{equation}
where $c_{MLE} > 0$ is a universal constant and $\mathcal{N}_{\mathcal{G}_\psi}(\epsilon)$ is the $\epsilon$-bracketing number of $\mathcal{G}_\psi$ \cite{provable24}.
Then we have ${P}(\mathcal{E}_a) \geq 1 - \delta/2$,
namely, with probability at least $1 - \delta / 2$ we have that for the GT reward function $\psi^{\star} \in \Psi(\mathcal{\zeta})$ .

\label{lemma1}
\end{lemma}

\begin{lemma}[Reward Estimation Error (In \cite{provable24}, Lemma 2)] \

% Let $\mathcal{E}_2$ be the event that the following inequality holds for any $\delta \in (0,1)$,

Under the event in Lemma \ref{lemma1}, 
with probability at least $1 - \delta / 2$, we have for all reward functions $\psi \in \mathcal{G}_\psi$ that
\begin{equation}
\mathbb{E}_{\tau^0 \sim \mu_0, \tau^1 \sim \mu_1} \left[ \left\| P_\psi(\cdot \mid \tau^0, \tau^1) - P_{\psi^\star}(\cdot \mid \tau^0, \tau^1) \right\|_1^2 \right] \leq \frac{c_{\text{TV}}}{N_p} \left( \sum_{n=1}^{N_p} \log \left( \frac{P_{\psi^\star}(o^n \mid \tau_n^0, \tau_n^1)}{P_\psi(o^n \mid \tau_n^0, \tau_n^1)} \right) + \log \left( \frac{\mathcal{N}_{\mathcal{G}_\psi}}{N_p\delta}  \right) \right)
\label{eq:e2}
\end{equation}
where $c_{\text{TV}} > 0$ is a universal constant.

Denote the event that Equation \ref{eq:e2} holds by $\mathcal{E}_b$ and then we know $P(\mathcal{E}_b) \geq 1 - \delta / 2$. Then from Lemma \ref{lemma1}, we know that under event $\mathcal{E}_1 = \mathcal{E}_a \cap \mathcal{E}_1$, we have for all $\psi \in \Psi(\mathcal{\zeta})$:
\begin{equation}
\mathbb{E}_{\tau^0 \sim \mu_0, \tau^1 \sim \mu_1} \left[ \left\| P_\psi(\cdot \mid \tau^0, \tau^1) - P_{\psi^\star}(\cdot \mid \tau^0, \tau^1) \right\|_1^2 \right] \leq \frac{c_g \log (\mathcal{N}_{\mathcal{G}_\psi}/ N_p\delta)}{N_p}
\label{eq:e21}
\end{equation}
where $c_g > 0$ is a universal constant.

Then under Assumption 3, we can apply the mean value theorem between $r^\star(\tau^1) - r^\star(\tau^0)$ and $r_\psi(\tau^1) - r_\psi(\tau^0)$ to Equation \ref{eq:e21} and ensure for all $\psi \in \Psi(\mathcal{\zeta})$ that
\begin{equation}
\mathbb{E}_{\tau^0 \sim \mu_0, \tau^1 \sim \mu_1} \left[ \left| \left( r_{\psi^\star}(\tau^1) - r_{\psi^\star}(\tau^0) \right) - \left( r_\psi(\tau^1) - r_\psi(\tau^0) \right) \right|^2 \right] \leq \frac{c \kappa^2 \log (\mathcal{N}_{\mathcal{G}_\psi} / N_p\delta)}{N_p}
\label{eq:e22}
\end{equation}
where $\kappa := \frac{1}{\inf_{x \in [-r_{\max}, r_{\max}]} \Phi'(x)}$ measures the non-linearity of the link function $\Phi$.

That is, under Assumption 3, $P(\mathcal{E}_1) \geq 1 - \delta$, namely, Equation \ref{eq:e22} holds with probability $1 - \delta$.
\label{lemma2}
\end{lemma}

\begin{lemma}[$\delta$-Quantifiers (In \cite{hu2023provable}, Lemma C.3)] \

Let $c > 0$ is an universal constant, $d$ is the length of the parameter vector $\psi$ , $\sigma \in (0, 1)$ is the confidence parameter and feature map $\phi : S \times A \rightarrow \mathbb{R}^d$, $N$ is the number of transitions in the candidate datasets, and
\begin{equation}
\Lambda = \nu I + \sum_{\tau=1}^{N}, \phi(s_{\tau}, a_{\tau}) \phi(s_{\tau}, a_{\tau})^\top \quad \beta = c \cdot d V_{\max} \sqrt{\xi}, \quad \xi = \log \left(\frac{2dN}{(1 - \gamma)\delta}\right)    
\end{equation}

Then $\Gamma = \beta \cdot \left(\phi(s, a)^\top \Lambda^{-1} \phi(s, a) \right)^{1/2}$ are $\delta$-quantifiers with probability at least $1 - \delta$. That is, let $\mathcal{E}_2$ be the event that the following inequality holds,
\begin{equation}
\left| (\mathbb{B}\widehat{V})(s, a) - (\widehat{\mathbb{B}}V)(s, a) \right| \leq \Gamma = \beta \sqrt{\phi(s, a)^\top \Lambda^{-1} \phi(s, a)}, \quad \forall (s, a) \in \mathcal{S} \times \mathcal{A}
\label{eq:eq3}
\end{equation}
\

Then we have ${P}(\mathcal{E}_2) \geq 1 - \delta$.
\label{lemma3}
\end{lemma}

Define the coverage coefficient $C^{\dagger}$ of a dataset $\mathcal{D} = \{(s_{i}, a_{i}, r_{i})\}_{i=1}^{N}$ as
\begin{equation}
C^{\dagger} = \sup_{C} \left\{ \frac{1}{N} \cdot \sum_{i=1}^{N} \phi(s_{i}, a_{i}) \phi(s_{i}, a_{i})^\top \succeq C \cdot \mathbb{E}_{\pi^\star} \left[ \phi(s_t, a_t) \phi(s_t, a_t)^\top \mid s_0 = s \right], \forall s \in \mathcal{S} \right\}
\end{equation}
The coverage coefficient $C^{\dagger}$ represents the maximum ratio between the density of empirical state-action distribution and the density induced from the optimal policy. Intuitively, it represents the quality of the dataset. For example, the \textit{expert} dataset has a high coverage ratio while the \textit{random} dataset may have a low ratio.
Then we have the following lemmas:

\begin{lemma}[Value Estimation Error (In \cite{hu2023provable}, Lemma C.1)] \

Under the event in Lemma \ref{lemma3}, we have
\begin{equation}
    V_{\psi}^{\pi^\star}(s) - \widehat{V}_{\psi}(s) \leq \frac{2cr_{\max}}{(1 - \gamma)^2} \sqrt{\frac{d^3 \xi}{C^{\dagger} N}}
\end{equation}
with probability $1 - \delta$ under Assumption \ref{assumption3}.
\label{lemma4}
\end{lemma}

\begin{lemma}[Pessimistic Value Estimate (In \cite{hu2023provable}, Lemma C.2)] \

Under the event in Lemma \ref{lemma3}, we have
\begin{equation}
\widehat{V}_{\psi}(s) - V_{\psi}^{\pi}(s) \leq 0
\end{equation}
with probability $1 - \delta$ under Assumption \ref{assumption3}.
\label{lemma5}
\end{lemma}

\subsubsection{A3. Theorem and Proof} \

\noindent \textbf{Theorem1 (Performance Bound)} \

\textit{Let} $\xi_\delta=\log \left(\frac{4d(N_pH_p+N_oH_o)}{(1-\gamma) \delta} \right)$ \textit{and} 
\begin{equation}
    C_\psi(\mathcal{G}_\psi, \pi^\star, \mu_{ref}) := \max \left\{ 0, \sup_{\psi \in \mathcal{G}_\psi} \frac{ \mathbb{E}_{\tau^0 \sim \pi^\star, \tau^1 \sim \mu_{ref}} \left[ r_{\psi^\star}(\tau^0) - r_{\psi^\star}(\tau^1) - r_\psi(\tau^0) + r_\psi(\tau^1) \right] } { \sqrt{ \mathbb{E}_{\tau^0 \sim \mu_0, \tau^1 \sim \mu_1} \left[ \left( r_{\psi^\star}(\tau^0) - r_{\psi^\star}(\tau^1) - r_\psi(\tau^0) + r_\psi(\tau^1) \right)^2 \right] } } \right\}
\label{eq:cr}
\end{equation}

\textit{For any} $\delta \in (0,1)$, \textit{under Assumption \ref{assumption1}, \ref{assumption2}, \ref{assumption3}, and if $\widehat{\pi}$ is any mesurable function of the data $\mathcal{D}_{pref}$, then with probability $1 - 2 \delta$, we have}

\begin{equation}
\begin{aligned}
\label{theoremx}
\text{SubOpt}(\widehat{\pi}; s) &=  V_{\psi^{\star}}^{\pi^{\star}}-V_{\psi^{\star}}^{\widehat{\pi}}  \\
&\le
\sqrt{\frac{c C_\psi^2(\mathcal{G}_\psi, \pi^{\star}, \mu_{ref}) \kappa^2 \log(\mathcal{N}_{\mathcal{G}_\psi} / N_p\delta)}{N_p}}
+
\frac{2c r_{\max}}{(1 - \gamma)^2} \sqrt{\frac{d^3 \xi_\delta}{N_p H_p C_p^\dagger + N_o H_o C_o^\dagger}}
\end{aligned}
\end{equation}

\begin{proof} 

Let $\mathcal{E}_1$ be the event $\psi^{\star} \in \Psi(\zeta)$, then we have $P(\mathcal{E}_1) \le 1-\delta$ from Lemma \ref{lemma2}; and
$\mathcal{E}_2$ be the event  where the Equation \ref{eq:eq3} holds, then we have $P(\mathcal{E}_2) \le 1-\delta$ from Lemma \ref{lemma3}.

Condition on $\mathcal{E}_1 \cap \mathcal{E}_2$, we have:
\begin{equation}
\begin{aligned}
V_{\psi^{\star}}^{\pi^{\star}}-V_{\psi^{\star}}^{\widehat{\pi}}  = &  (V_{\psi^{\star}}^{\pi^{\star}} - V_{\widetilde{\psi}}^{\pi^{\star}}) 
- (\widehat{V}_{\psi^{\star}} - \widehat{V}_{\widetilde{\psi}}) 
+ (V_{\widetilde{\psi}}^{\pi^\star} - \widehat{V}_{\widetilde{\psi}})
+ \underbrace{(\widehat{V}_{\psi^{\star}} - V_{\psi^{\star}}^{\widehat{\pi}})}_{\le 0, \text{Lemma \ref{lemma5}}} \\
 \le & (V_{\psi^{\star}}^{\pi^{\star}} - V_{\widetilde{\psi}}^{\pi^{\star}}) 
- (\widehat{V}_{\psi^{\star}} - \widehat{V}_{\widetilde{\psi}}) 
+ (V_{\widetilde{\psi}}^{\pi^\star} - \widehat{V}_{\widetilde{\psi}}) \\
 = & \left((V_{\psi^{\star}}^{\pi^{\star}}-\mathbb{E}_{\tau \sim \mu_{ref}}[r_{\psi^\star}(\tau)]) - (V_{\widetilde{\psi}}^{\pi^{\star}}-\mathbb{E}_{\tau \sim \mu_{ref}}[r_{\widetilde{\psi}}(\tau)])\right) \\
& - \underbrace{ \left((\widehat{V}_{\psi^{\star}}-\mathbb{E}_{\tau \sim \mu_{ref}}[r_{\psi^\star}(\tau)]) - (\widehat{V}_{\widetilde{\psi}}-\mathbb{E}_{\tau \sim \mu_{ref}}[r_{\widetilde{\psi}}(\tau)]) \right) }_{\ge 0, \text{Pessimistic Reward Estimation}} 
+ (V_{\widetilde{\psi}}^{\pi^\star} - \widehat{V}_{\widetilde{\psi}}) \\
\le &  \left((V_{\psi^{\star}}^{\pi^{\star}}-\mathbb{E}_{\tau \sim \mu_{ref}}[r_{\psi^\star}(\tau)]) - (V_{\widetilde{\psi}}^{\pi^{\star}}-\mathbb{E}_{\tau \sim \mu_{ref}}[r_{\widetilde{\psi}}(\tau)])\right)
+ (V_{\widetilde{\psi}}^{\pi^\star} - \widehat{V}_{\widetilde{\psi}}) \\
= & \underbrace{ \mathbb{E}_{\tau^0 \sim \pi^{\star}, \tau^1 \sim \mu_{ref}} \left[ (r_{\psi^\star}(\tau^0) - r_{\psi^\star}(\tau^1)) - (r_{\widetilde{\psi}}(\tau^0) - r_{\widetilde{\psi}}(\tau^1)) \right] }_{ \text{Lemma \ref{lemma2}} }
+  (V_{\widetilde{\psi}}^{\pi^\star} - \widehat{V}_{\widetilde{\psi}}) \\
\leq & \underbrace{ C_\psi(G_r, \pi^{\star}, \mu_{ref}) \sqrt{\mathbb{E}_{\tau^0 \sim \mu_0, \tau^1 \sim \mu_1} \left[ |r_{\psi^\star}(\tau^0) - r_{\psi^\star}(\tau^1) - r_{\widetilde{\psi}}(\tau^0) + r_{\widetilde{\psi}}(\tau^1)|^2 \right]} }_{\text{Definition of $C_\psi$, Equation \ref{eq:cr}} }
+ \underbrace{ (V_{\widetilde{\psi}}^{\pi^\star} - \widehat{V}_{\widetilde{\psi}}) }_{\text{Lemma \ref{lemma4}}}\\
\le & \sqrt{\frac{c C_\psi^2(\mathcal{G}_\psi, \pi^{\star}, \mu_{ref}) \kappa^2 \log(\mathcal{N}_{\mathcal{G}_\psi} / N_p\delta)}{N_p}}
+
\frac{2c r_{\max}}{(1 - \gamma)^2} \sqrt{\frac{d^3 \xi_\delta}{N_p H_p C_p^\dagger + N_o H_o C_o^\dagger}}
\end{aligned}
\end{equation}

From the union bound, we have that the above inequality holds with a probability of $1-2\delta$.
\end{proof}

%Unlike \cite{hu2023provable}, which assumes ground-truth and reward-free data, and \cite{provable24}, which assumes an optimal downstream reward in offline PbRL, we consider both reward learning bias (first term in Equation \ref{eq:subopt}) and errors in offline RL under limited data (second term in Equation \ref{eq:subopt}), leading to a theoretical upper bound for offline PbRL.

Compare to \cite{hu2023provable}, which establishes a performance bound for exploiting reward-free data in offline RL, we extend this result by proving a bound for leveraging out-dataset trajectories in 
 to  trajectory-based comparison offline PbRL. 
 The first term in Equation \ref{theoremx} suggests that increasing  $N_p$  can enhance theoretical performance, as a larger dataset facilitates more accurate training of the reward model.
 Compare to \cite{provable24}, which assumes that downstream RL algorithms can attain the optimal policy, we consider the value function that needs to be estimated. 
 The second term in Equation \ref{theoremx} demonstrates that increasing the amount of offline data helps to reduce the error in value function estimation for offline RL.

% \begin{corollary}[Performance Bound: Tighter Bounds] \

% For any $\delta \in (0,1)$, under Assumption \ref{assumption1}, \ref{assumption2}, and \ref{assumption3}, with probability $1 - 2 \delta$, we have

% \begin{equation}
% \begin{aligned}
% \text{SubOpt}(\widehat{\pi}; s) 
% \le&
% \mathbb{I}_{(\pi^{\star} \in \mathcal{D}_p)}  \sqrt{\frac{c C_\psi^2(\mathcal{G}_\psi, \pi^{\star}, \mu_{\text{ref}}, \mathbb{I}_{(\pi^{\star} \in \mathcal{D}_p)}) \kappa^2 \log(\mathcal{N}_{\mathcal{G}_\psi}/ N\delta)}{N}} \\
% &+
% \mathbb{I}_{(\pi^{\star} \notin \mathcal{D}_p)} \sqrt{\frac{c C_\psi^2(\mathcal{G}_\psi, \pi^{\star}, \mu_{\text{ref}}, \mathbb{I}_{(\pi^{\star} \notin \mathcal{D}_p)}) \kappa^2 \log(\mathcal{N}_{\mathcal{G}_\psi} / N\delta)}{N}} \\
% &+
% \frac{2c r_{\max}}{(1 - \gamma)^2} \sqrt{\frac{d^3 \xi_\delta}{N_p H_p C_p^\dagger + N_o H_o C_o^\dagger}}
% \end{aligned}
% \end{equation}
% \end{corollary}

\clearpage
\section{B. More Related Works and Clarifications}
% 在这个章节，我们提供更多相关工作以便读者全面了解我们的方法基础；并且针对审稿中存在的普遍意见（尤其是在新奇性和与offline RL算法的细节的对比方面）进行进一步澄清。
In this section, we provide additional related works to help readers gain a comprehensive understanding of the foundations of our approach. 
\textbf{Furthermore, we enhance clarity on common concerns raised during the review process, particularly regarding novelty and detailed comparisons with offline RL algorithms.}

\subsubsection{B1. Rubost Reward Modeling} \

Various works aim to enhance the credit assignment in reward models for trajectory rewards by employing more robust reward modeling techniques. 
\cite{non-markov22} suggests using non-Markov models to capture temporal dependencies in labeled trajectories. 
PT \cite{pt23} posits that humans are highly sensitive to significant moments and uses transformers for reward modeling, explicitly learning the reward weight for each step via a bidirectional preference attention layer. 
Similarly, PRIOR \cite{prior} estimates state importance based on hindsight information from world models.

On the other hand, several studies focus on modeling reward functions rather than directly using the Bradley-Terry model.
For example, CPL \cite{cpl24} employs a regret-based model of preferences instead of the widely accepted partial return model, which only considers the sum of rewards, as the regret-based model directly informs the optimal policy. 
RIME \cite{rime24} introduces an additional classifier to identify erroneous noise, enabling the reward model to recognize inaccurate human labels. 
HPL \cite{hpl24} takes advantage of the unlabeled dataset by learning a prior over future outcomes, thus incorporating trajectory distribution information from the unlabeled dataset.

Moreover, appropriate data and model augmentation can improve the generalization of reward models. 
For instance, increasing the amount of preference data through clipping \cite{ftb24,qpa24} or enhancing the multimodal nature of reward models via model ensemble techniques \cite{oprl23}.

\subsubsection{B2. Improving Stitching Ability of CSM} \

The trajectory stitching capability of CSM remains preliminary, partly due to the direct application of the Transformer to decision tasks without modification \cite{tfmsurvey23} . 
Thus, it is essential to adapt the Transformer to the specific characteristics of decision tasks or the multi-modality of trajectories. 
Some studies try to classify or preprocess the input sequences.
For example, StARformer \cite{starformer22} employs a Step Transformer Layer to capture local information and a Sequence Transformer Layer to capture overall sequence information. 
RADT \cite{radt24} decouples rewards from state-action pairs explicitly through cross-attention. 
GCPC \cite{gcpc24} extracts the effective information of target trajectories using a bidirectional Transformer, making it efficient for downstream policies. 
Additionally, in terms of structure, DC \cite{dc24} proposes replacing the Attention module with local convolutions, achieving better performance with fewer parameters.
% , while DeMa \cite{dai2024mamba} employs the more advanced Mamba architecture to capture long-term dependencies.

An alternative approach involves integrating TD-Learning to enhance trajectory stitching capabilities, which aligns with our concept. 
For instance, QDT \cite{qdt23} relabels trajectory returns using a value function, thereby improving performance with limited optimal data. 
EDT \cite{edt23} dynamically adjusts the history length via a value function to optimize trajectories for higher returns.
CGDT \cite{cgdt22} explicitly constrains policy outputs to align with target $Q$ values, rather than selecting actions with the highest returns in the dataset.
$Q$-Transformer \cite{qtransformer23} and QT \cite{qt24} jointly maximize the $Q$ value to guide action selection in DT outputs.

\subsubsection{B3. Clarifications on Novelty} \

Some reviewers may perceive the results in this paper as merely a clever integration of past approaches. In response, we would like to emphasize the core contributions of our work:

Our core contribution lies in finding that \textbf{DT based in-dataset regularization tackles the challenge of inaccurate TD-learning stitching caused by reward bias for offline PbRL.} 
By combining DT and TDL dynamically, alongside an ensemble normalization to address reward inaccuracies, DTR outperforms the complex diffusion-based SOTA method FTB \cite{ftb24} significantly. We believe that DTR offers a high and easy-to-implement baseline for offline PbRL. 

\subsubsection{B4. Clarifications on Technical Details} \

Some reviewers may question the clarity of using a decision transformer; therefore, we have reiterated the differences between in-dataset regularization and traditional offline RL policy regularization, clarifying our technical advantages:

Traditional policy regularization techniques in offline RL (e.g., TD3BC and IQL) primarily focus on constraining policy actions to remain close to the behavior policy or mitigating value overestimation through Q-function modifications \textbf{based on step-level regularization}. 
However, these methods \textbf{do not fully address the issue of trajectory stitching errors that arise from overestimations induced by inaccurate reward models}, which are particularly pronounced in offline PbRL. 
In offline PbRL, \textbf{DT-based trajectory-level regularization} offers several advantages for our specific context:

1.DT models entire trajectories through return-to-go (RTG) conditioning, \textbf{aligning with in-dataset trajectories of high returns and avoiding overoptimistic trajectory stitching of traditional offline RL}.

2.Unlike step-level regularization, trajectory-level regularization \textbf{incorporates multi-step trajectory information, balancing fidelity to high-return paths with exploration, reducing the risk of value overestimation due to reward bias}.

For experimental results, we can see \textbf{traditional step-level regularization techniques can not perform well in offline PbRL}. 
As shown in Table \ref{tab:mujoco}, Pb-IQL (score: 73.29) and Pb-TD3BC (score: 51.86) are both worse than DTR (score: 82.88). This conclusion is also consistent on the Adroit manipulation benchmark as shown in Table \ref{tab:adroit}.

\clearpage
\section{C. Environments and Dataset Details}
\begin{figure}[h!]
\centering
 \includegraphics[width=.7\linewidth]{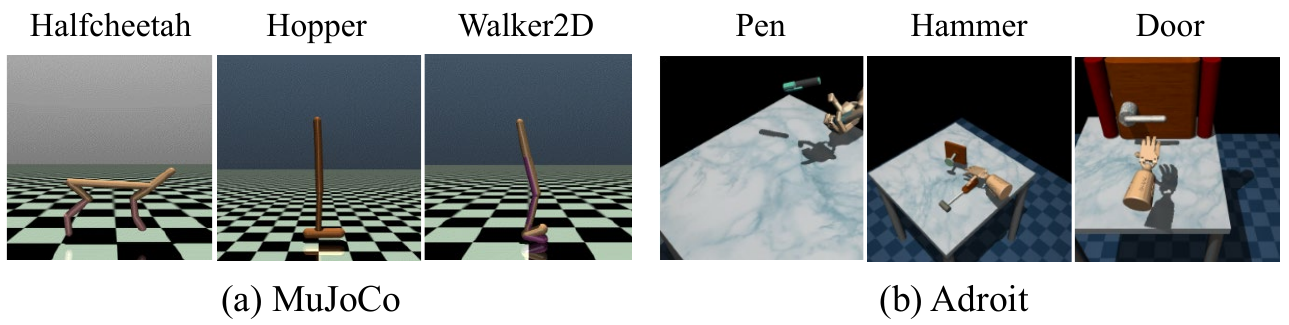}
 \caption{Visualization of each task in the environments. }
 \label{fig:env}
\end{figure}

We provide a detailed description of the experimental environment and dataset and visualization of each task in Figure \ref{fig:env}.
Detailed information about the trajectory number of the offline datasets, the preference query number, and the query length of preference datasets are shown in Table \ref{tab:dataset_details}.

\subsubsection{C1. MuJoCo}
MuJoCo \cite{openaigym} environment are widely used benchmarks in previous research on offline RL.
It contains three domains of continuous control:
Walker2d, which is a control task of a planar walker, which aims at moving as fast as possible while keeping balance;
Hopper, which is a control task involving a single-legged robot where the goal is to get the robot to jump as rapidly as possible;
HalfCheetah, which is a control task that aims to make a simulated robot perform a cheetah-like running motion with the goal of maximizing the distance traveled within a given time frame.
D4RL \cite{d4rl20} includes mixed datasets to evaluate the effects of heterogeneous policy combinations. 
The "medium" (m) dataset consists of 1 million transitions sampled from trajectories produced by a soft actor-critic agent. 
The "medium-replay" (m-r) dataset includes all samples stored in the replay buffer during training until the policy attains a "medium" level of performance. 
The "medium-expert" (m-e) dataset combines equal parts of expert demonstrations and suboptimal data, generated either by a partially trained policy or by running a uniformly random policy.

\subsubsection{C2. Adroit}
The Adroit manipulation platform \cite{Kumar2016thesis} involves controlling a 24-DoF robotic hand. There are 3 tasks in our experiments: pen (aligning a pen with a target orientation), door (opening a door), and hammer (hammering a nail into a board).
These tasks evaluate the impact of narrow expert data distributions and human demonstrations on some sparse reward, high-dimensional robotic manipulation tasks.
For each task, it includes three types of datasets: “human”, which consists of a small amount of demonstration data from humans with 25 trajectories per task, and “cloned”, which is a small amount of additional data from a policy trained on the demonstrations.

\begin{table}[h!]
\centering
\begin{tabular}{clcccccc}
\toprule
\textbf{Domain} & \textbf{Env Name} &  \textbf{Obs Dim} &  \textbf{Action Dim}& \textbf{Trajectories Num} & \textbf{Queries Num} & \textbf{Queries Length}\\
\midrule

\multirow{9}{*}{Gym MuJoCo} & Walker2d-m & 17 & 6& 1190 & 2000 & 200  \\
                            & Walker2d-m-r &17 &6 & 890& 2000 & 200  \\
                            & Walker2d-m-e &17 &6 & 2190 & 2000 & 200  \\
                            & Hopper-m &11 &3 & 2186 & 2000 & 200  \\
                            & Hopper-m-r &11 &3 & 1705 & 2000 & 200  \\
                            & Hopper-m-e &11 & 3& 3213 & 2000 & 200  \\
                            & Halfcheetah-m &17 &6 & 999& 2000 & 200 \\
                            & Halfcheetah-m-r &17 &6 & 201 & 2000 & 200  \\
                            & Halfcheetah-m-e &17 &6 & 1999 & 2000 & 200  \\ 
\midrule
\multirow{6}{*}{Adroit}     & Pen-human &45 &24 & 25 & 2000 & 50  \\
                            & Pen-cloned  &45 &24 & 4357 & 2000 & 50  \\
                            & Door-human &39 &28 & 25 & 2000 & 50  \\
                            & Door-cloned &39 &28 & 3605 & 2000 & 50 \\
                            & Hammer-human &46 &26 & 25 & 2000 & 50 \\
                            & Hammer-cloned &46 &26 & 3754 & 2000 & 50  \\
\bottomrule
\end{tabular}
\caption{Datasets details.}
\label{tab:dataset_details}
\end{table}

\clearpage

\section{D. Implementation Details}
\subsubsection{D1. Computing Resources} \

We trained DTR on an RTX3090 GPU, and all code was written in PyTorch. Specifically, on the MuJoco task, it takes us about two minutes to train the reward model and about 10 hours to train and evaluate the policy. And policy training requires about 2.7 GB of GPU memory (when $H$=20). In contrast, Pb-IQL training and policy evaluation take about 5 hours and about 2.2 GB of GPU memory, which is lower than DTR. The reason is that DTR also needs to generate trajectories through Transformer while training the $Q$ function, which consumes extra time. Compared with the huge consumption of 36 h/23 GB of FTB (from Appendix B1 of \cite{ftb24}), DTR has greater time and GPU computing resource advantages.

\subsubsection{D2. Reward Model Details} \

In DTR, we use an MLP-based reward model, the according hyperparameters are recorded in Table \ref{tab:hyperparameters_reward}.
As suggested by \cite{uni-rlhf}, we train for 100 epochs. To prevent overfitting, we also test the accuracy of reward evaluation on the training set. When the accuracy is greater than 96.8\%, we stop training and save the last checkpoint to mark the reward-free dataset.
In addition, some works use transformer-based reward modeling \cite{pt23,ftb24}. However, we found that the transformer-based reward model suffers from extrapolation errors due to the inconsistency between the context lengths of train and inference  (see the section E2 for further discussion).

\begin{table}[h!]
\centering
\begin{tabular}{ll}
\toprule
% \textbf{Reward Model} & \\
% \midrule
\textbf{Hyperparameter} & \textbf{Value} \\
\midrule
Structure & MLP \\
Ensemble Size & 3 \\
Batch Size & 64 \\
Training Epochs & 100 \\
Hidden Size & 256 \\
Network Layers & 3 \\
Learning Rate & 3e-4 \\
Activation Function & ReLU \\
Output Activation & Tanh \\
\bottomrule
\end{tabular}
\caption{Hyperparameters for the Reward Model.}
\label{tab:hyperparameters_reward}
\end{table}

\subsubsection{D3. Downstream Model Details} \

For the downstream offline RL model, the according hyperparameters are in Table \ref{tab:hyperparameters_qt}. Specifically, we follow the DT \cite{dt21} settings, training the Transformer for 1000 steps per epoch, evaluating with 10 trajectory rollouts after each epoch, for up to 50 epochs.
For the $Q$ network part, we follow the settings of TD3BC \cite{TD3BC21} and add state normalization and learning rate decay.
For the context length $H$ and the maximum coefficient $\eta_{\max}$, adjustments need to be made in different environments to achieve the best performance. The relevant ablation experiments will be analyzed in the next section.

\begin{table}[h!]
\centering
\begin{tabular}{ll|ll}
\toprule
\textbf{Hyperparameter} & \textbf{Value} & \textbf{Hyperparameter} & \textbf{Value} \\
\midrule
Batch Size & 256 & Context Len $H$ for MuJoCo & 5 for Halfcheetah-m, Halfcheetah-m-r \\
Embedding Size & 256 & & 20 otherwise \\
Number of Layers & 4 & Context Len $H$ for Adroit & 20 for all tasks \\
Number of Attention Heads & 4 & $\eta_{\max}$ for Halfcheetah & 1.0 for medium, medium-replay \\
Activation Function & ReLU & & 0.1 for medium-expert \\
Learning Rate & 3e-4 & $\eta_{\max}$ for Hopper & 1.0 for medium, medium-expert \\
Dropout & 0.1 & & 3.0 for medium-replay \\
Number of Steps Per Iteration & 1000 & $\eta_{\max}$ for Walker2D & 2.0 for all datasets \\
Maximum Number of Iterations & 50 & $\eta_{\max}$ for Pen & 0.1 for all datasets \\
Discount Factor  $\gamma$ & 0.99 & $\eta_{\max}$ for Hammer & 0.01 for clone, 0.1 for human \\
EMA Factor $\tau$ & 5e-3 & $\eta_{\max}$ for Door & 0.001 for clone, 0.005 for human \\
Weight Decay Factor & 1e-4 & & \\
Learning Rate Decay & True & & \\
State Normalization & True & & \\
Number of Eval Episodes & 10 & & \\
\bottomrule
\end{tabular}
\caption{Hyperparameters for Downstream Offline RL Model.}
\label{tab:hyperparameters_qt}
\end{table}

\subsubsection{D4. Pseudocode} \

Algorithm 1 shows the entire pseudocode of DTR, which corresponds to the process in Figure \ref{fig:main}.

\begin{algorithm}
\caption{DTR}
\begin{algorithmic}[1]
\REQUIRE Preference dataset $\mathcal{D}_{pref}$; Reward-free offline datasets $\mathcal{D}$; Ensemble size $N$;  Maximum $Q$ coefficient $\eta_{\max}$; Sequence horizon $H$; EMA coefficient $\rho$.
\STATE {{\textbf{\textit{\#  (1) Learn Reward Model and Relabel Offline Dataset.}}}}
\STATE Initialize ensemble reward model $\{\hat{r}_{\psi_n}\}_{n=1}^{N}$.
\STATE \textbf{For} $\text{Step} = 1$ to $M$ \textbf{do}:
\STATE \quad  Sample mini-batch $\mathcal{B}_{pref} = \{(\sigma^0, \sigma^1, y)\} \sim \mathcal{D}_{pref}$.
\STATE \quad  Update $\{\hat{r}_{\psi_n}\}_{n=1}^{N}$ by minimizing Equation \ref{eqt:pref_ce}.
% \STATE  \textbf{End for}
\STATE \textbf{End For}
\STATE Relabel offline dataset with $\{\hat{r}_{\psi_n}\}_{n=1}^{N}$ and ensemble normalization by Equation \ref{eqt:reward_norm}.
\STATE Calculate $\hat{R}_i$ for each state $s_i$, then get $\mathcal{D}$ with $\hat{R}$.

\STATE
\STATE {{\textbf{\textit{\# (2) Training Policy and Q-Network.}}}}
\STATE Initialize policy network $\pi_\theta$; critic networks $Q_{\phi_1}$, $Q_{\phi_2}$; target networks $\pi_{\theta'}$, $Q_{\phi_1'}$, $Q_{\phi_2'}$; $Q$ coefficient $\eta_{\text{init}}=0$.

\STATE \textbf{For} $\text{Step} = 1$ to $T$ \textbf{do}:
    \STATE \quad Sample batch $\mathcal{B} = \{(\hat{R}_j, s_j, a_j, \hat{r}_j)_{j=t}^{t+H-1}\} \sim \mathcal{D}$.
    % \STATE \quad // Q-value function learning
    \STATE \quad Sample $\hat{a}_{t+H} \sim \pi_{\theta}(\hat{R}_{t:t+H}, s_{t:t+H}, a_{t:t+H-1})$.
    \STATE \quad  Update $Q_{\phi_1}$ and $Q_{\phi_2}$ by Equation \ref{eqt:n_step_Q}.
    % \STATE // Policy learning
    % \For {$i = 1$ to $K$}
    \STATE \quad  AR Sample  $\{\hat{a}_{t+i}\}_{i=1}^{H} \sim \pi_\theta(\hat{R}_{t:t+i}, s_{t:t+i}, a_{t:t+i-1})$.
    % \EndFor
    \STATE  \quad Update policy by minimizing Equation \ref{eqt:loss}.
    \STATE \quad $\theta' = \rho\theta + (1 - \rho)\theta'$, $\{\phi_i' = \rho\phi_i + (1 - \rho)\phi_i'\}_{i=1}^2$.
    \STATE \quad Update coefficient $\eta_{\text{step}}$ by Equation \ref{eqt:para}.
\STATE \textbf{End For}
\STATE
\STATE \# \textbf{\textit{ (3) Inference}}
\STATE Initialize the maximum RTG and targets $\{\hat{R}_i\}_{i=1}^K$, a trained policy $\pi_\theta$ and $Q$ network $Q_\phi$.
\STATE \textbf{Repeat}
    \STATE \quad Sample multiple actions with different RTGs: \\
    \quad \quad $\hat{a}_t^i = \pi_\theta(\hat{a}_t^i|\hat{R}_{t-H+1:t}, s_{t-H+1:t}, a_{t-H+1:t-1})$
    \STATE \quad Compute $Q$ value with $\{(s_t, \hat{a}_t^i)\}_{i=1}^K$.
    \STATE \quad Sample $a_t$ from $\{\hat{a}_t^i\}_{i=1}^m$ with the max $Q$ value.
    \STATE \quad Execute the action $a_t$ and collect the next state $s_{t+1}$.
    \STATE \quad Update current return-to-go $\hat{R}_{t+1}^i$ by Equation \ref{eqt:r_update}. 
\STATE \textbf{End}
\label{ref:algorithm}
\end{algorithmic}
\end{algorithm}

% \subsubsection{D5. Code Source} \

% Code source can be available at: {https://github.com/TU2021/DTR}

\clearpage
\section{E. Additional Experimental Results}

\subsubsection{E1. Complete Training Curve in MuJoCo} \

\begin{figure}[h]
 \centering
 \includegraphics[width=1\linewidth]{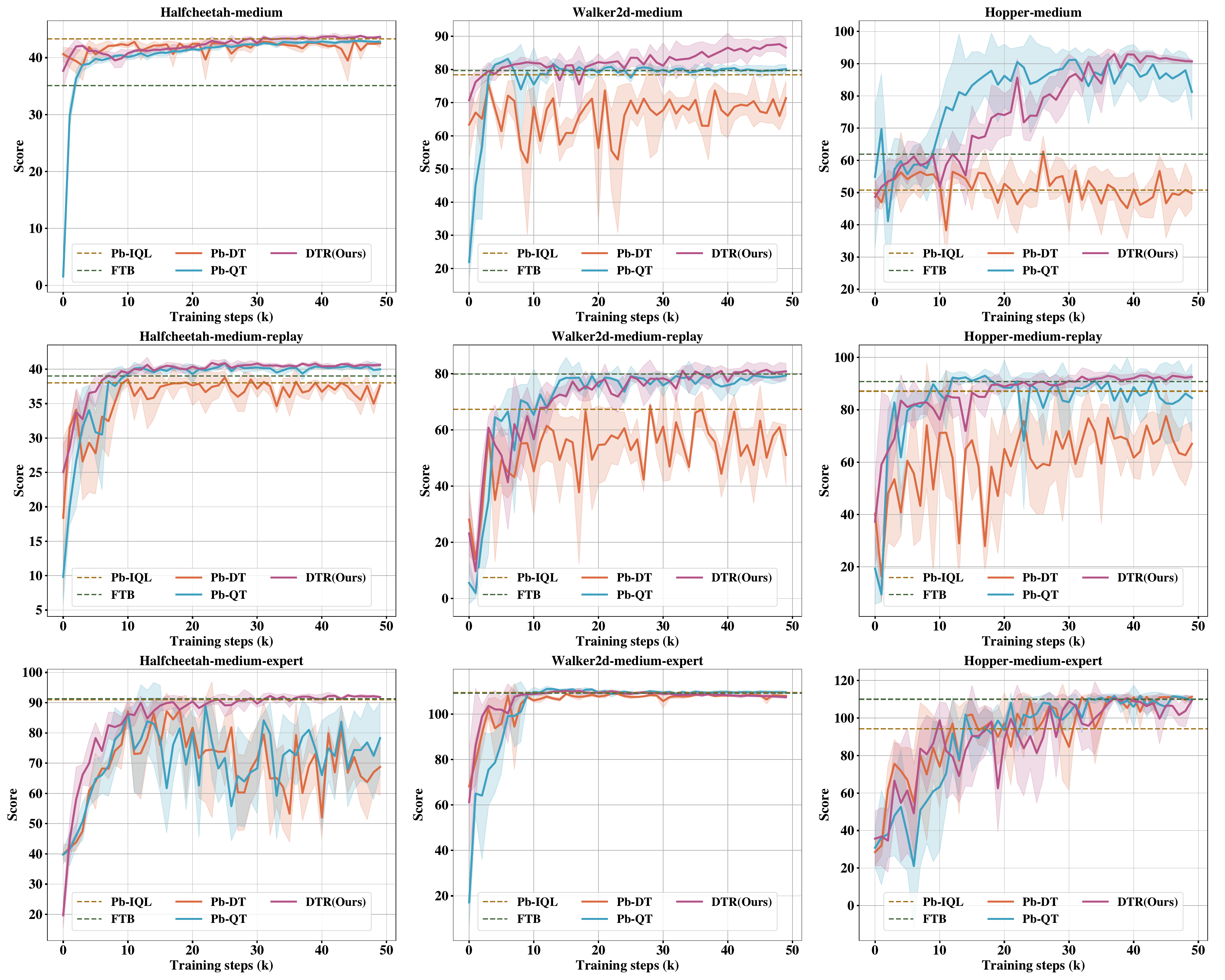}
 \caption{Complete Training Curve in MuJoCo. The two dotted lines represent the final evaluation scores of baseline Pb-IQL and FTB, respectively.}
 \label{fig:plot_exp}
\end{figure}

We provide the complete training curve of MuJoco in Figure \ref{fig:plot_exp} to facilitate the comparison of the training efficiency and convergence performance of different algorithms.
In contrast, DTR achieves SOTA performance in most environments, consistent with the records in Table \ref{tab:mujoco}.

\subsubsection{E2. Architecture of Reward Model : MLP vs Transformer} \

\begin{figure}[h]
 \centering
 \includegraphics[width=.85\linewidth]{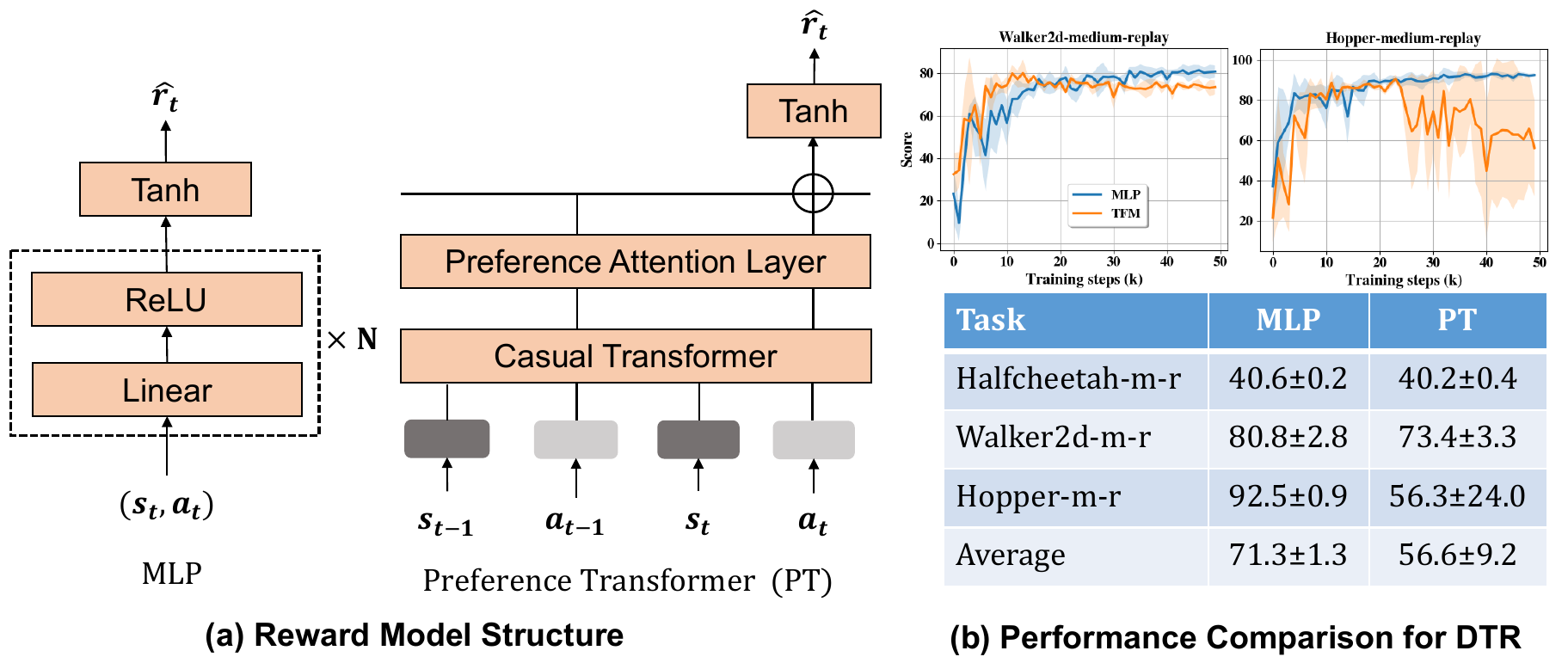}
 \caption{Comparison of reward model structures between MLP and PT, and performance comparison on downstream tasks of MuJoco-Medium-Replay.}
 \label{fig:reward_model}
\end{figure}

Some studies employ transformers to model reward functions. For instance, the Preference Transformer (PT) uses causal transformers to handle non-Markov rewards and explicitly calculates importance weights via a preference attention layer \cite{pt23}. A structural comparison between MLP and PT is presented in Figure \ref{fig:reward_model}a.
The goal of PT is to enhance the assignment of credit to trajectory rewards. However, our experiments indicate that PT-based rewards perform inferiorly compared to those generated by MLP. Additionally, PT exhibits significant training fluctuations, as shown in Figure \ref{fig:reward_model}b.

\begin{figure}[h]
 \centering
 \includegraphics[width=.8\linewidth]{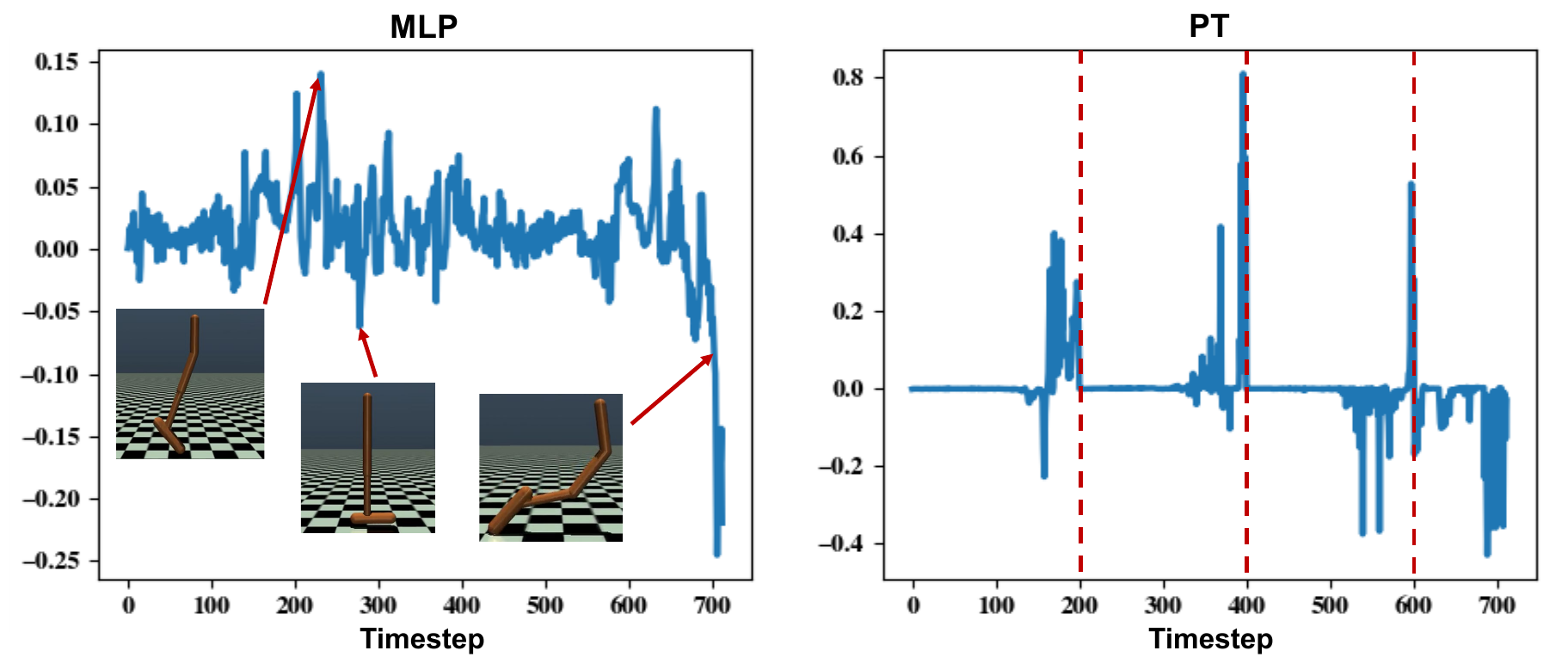}
 \caption{A hopper agent with an episode length of 712 steps, with reward changes under MLP and PT annotations.}
 \label{fig:reward_model_compare}
\end{figure}

Furthermore, in order to find out the reason for this phenomenon, we specifically plotted in Figure \ref{fig:reward_model_compare} the reward change curve of an episode with a length of 712 steps from the beginning of the jump to the final fall in the Hopper environment.
The MLP reward is a Markov reward, and the reward for the current state-action pair is only related to the current moment, while the PT is a non-Markov reward, and the reward at the current moment depends on the previous input.
We find that the MLP reward can well describe the reward changes during Hopper's jumping, landing, and falling, while the PT reward is not significant for these state changes, and in the first 600 steps, the reward value will change with a period of 200 steps.

We think the reason may be that the trajectory length of PT's training dataset is 200, and when annotating rewards, we often need to annotate trajectories with a length greater than 200 (for example, the maximum length of MuJoco is 1000).
Directly annotating the entire trajectory will lead to the length extrapolation problem; that is, the transformer will be confused by the unseen position encoding, causing the annotation to fail.
One solution is to cut the overly long trajectory into a length that PT can successfully annotate; for example, split a 1000-length trajectory into five 200-length trajectories, which leads to the periodicity of reward.
Similar issues were discussed in (https://openreview.net/forum?id=Peot1SFDX0), and our experiments verified the limitations of the transformer-based reward model under length extrapolation.
How to further improve the reward model architecture in the future will be an interesting research topic.

% \subsubsection{E3. More Ablation of Reward Normalization} \
% \begin{figure}[tb]
%  \centering
%  \includegraphics[width=.9\linewidth]{img/plot_reward_both_apendix.pdf}
%  \caption{The return values given by different normalization methods. Each dot represents one trial, with its coordinates indicating the estimated return values.}
%  \label{fig:plot_reward_both_apendix}
% \end{figure}

\subsubsection{E3. Ablation of Context Length $H$} \

\begin{figure}[h]
 \centering
 \includegraphics[width=\linewidth]{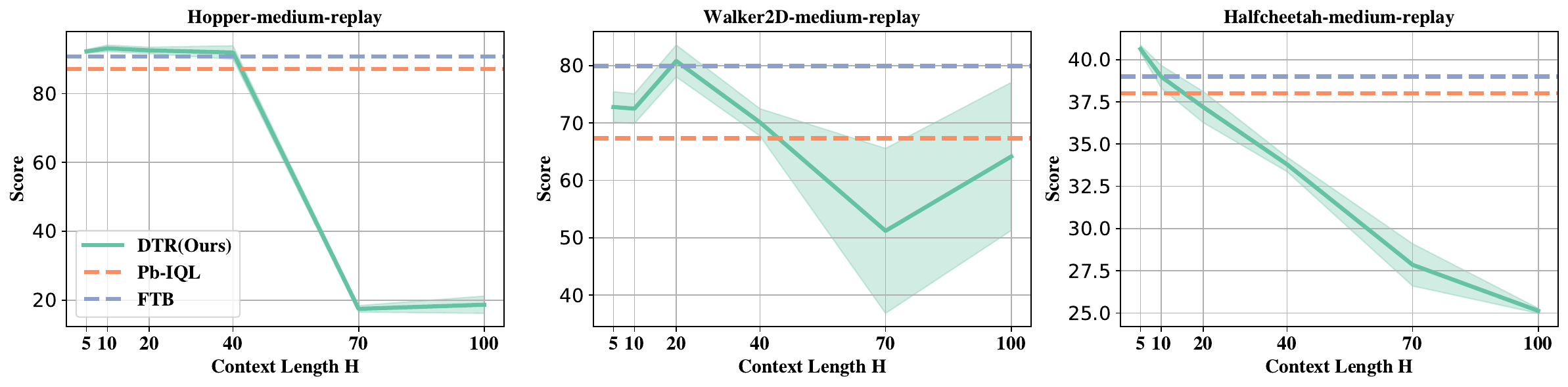}
 \caption{Performance comparison of different $H$ in MuJoCo-medium-replay tasks. In Hopper and Walker2d, $H=20$ is the optimal result, and in Halfcheetah, $H=5$.}
 \label{fig:plot_abl_len}
\end{figure}

In DT, context length $H$ is an important parameter that represents the Long Task Horizon Ability of the model. 
Although the decision at the current moment depends only on the current state in MDP, the  experiments of DT  show that past information can be valuable for sequence modeling methods in certain settings, where longer sequences tend to produce better results \cite{dt21}.
Figure \ref{fig:plot_abl_len} shows the performance comparison of different $H$ in the MuJoCo-medium-replay tasks. 
% Detailed evaluation scores and standard deviations are recorded in the Table \ref{tab:context_len_performance}
The results show that moderate $H$ can usually achieve better performance. 
For example, in Walker2D-medium-replay, $H=20$ achieves the best performance. 
A smaller $H$ may cause the transformer to be unable to model historical trajectory information, causing the algorithm to degenerate into TD3BC; a larger $H$ introduces redundant information, and a large step size introduces greater variance of $Q$ estimation, resulting in performance deterioration.

% \begin{table}[h]
% \centering
% \begin{tabular}{cccc}
% \toprule
% \textbf{$H$} & \textbf{Hopper-m-r}  & \textbf{Walker-m-r} & \textbf{Halfcheetah-m-r} \\
% \midrule
% 5   & \textbf{92.23} ± {\scriptsize 0.55}  & 72.77 ± {\scriptsize 2.66}  & \textbf{40.63*} ± {\scriptsize 0.46} \\
% 10  & \textbf{93.13*} ± {\scriptsize 0.94}  & 72.48 ± {\scriptsize 2.59}  & 38.98 ± {\scriptsize 0.67} \\
% 20  & \textbf{92.53} ± {\scriptsize 1.86}  & \textbf{80.79*} ± {\scriptsize 5.55}  & 37.19 ± {\scriptsize 0.92} \\
% 40  & 91.91 ± {\scriptsize 1.96}  & 70.09 ± {\scriptsize 2.38}  & 33.8 ± {\scriptsize 0.43} \\
% 70  & 17.45 ± {\scriptsize 0.94}  & 51.18 ± {\scriptsize 14.37}  & 27.85 ± {\scriptsize 1.25} \\
% 100 & 18.65 ± {\scriptsize 2.5}   & 64.14 ± {\scriptsize 12.9}   & 25.13 ± {\scriptsize 0.15} \\
% \bottomrule
% \end{tabular}
% \caption{Performance scores for different context lengths $H$.}
% \label{tab:context_len_performance}
% \end{table}

\subsubsection{E4. Ablation of Coefficient $\eta_{\max}$} \

\begin{figure}[h]
 \centering
 \includegraphics[width=\linewidth]{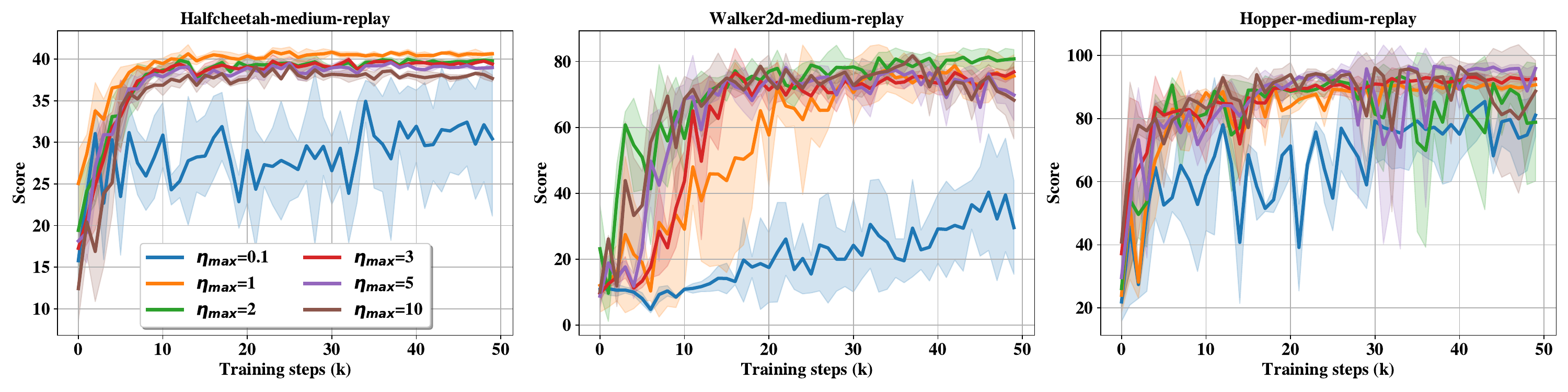}
 \caption{Training curve of different $\eta_{\max}$ in MuJoCo-medium-replay tasks. Considering the training stability and convergence performance, we take $\eta_{\max}=1$ in Halfcheetah, $\eta_{\max}=2$ in Walker2d and $\eta_{\max}=3$ in Hopper as the optimal coefficient.}
 \label{fig:plot_abl_eta}
\end{figure}

In practice, $\eta_{\max}$ is an important coefficient that balances in-dataset conservatism and out-dataset exploration. 
A high $\eta_{\max}$ will make the $Q$ loss ratio too large, which weakens the regularization term of the policy loss. This may lead to an overestimation of the $Q$ value of OOD state-action pairs, which is fatal in offline RL. $\eta_{\max}=+\infty$ will cause DTR to degenerate into Pb-TD3. 
A low $\eta_{\max}$ will make the policy too conservative and unable to bring into play the trajectory stitching ability of the $Q$ network module. $\eta_{\max}=0$ will cause DTR to degenerate into Pb-DT. 
The training curve in Figure \ref{fig:plot_abl_eta} shows that taking a compromise $\eta_{\max}$ can achieve optimal performance.

\subsubsection{E5. Details of Overall Ablation Performance.} \

In the main text, to investigate the effects of dynamic coefficients and reward normalization, we report the overall performance of ablation variants in Table 3. Detailed results for each task are provided in Table \ref{tab:gt2}.

\begin{table}[h]
\centering
% \begin{adjustbox}{width=\textwidth}
\begin{tabular}{lcccc|c}
\toprule
    & \textbf{Oracle($\hat{r}$)} & \textbf{+dynamic} $\eta$  & \textbf{+reward\_norm} & \textbf{ +dynamic} $\eta$ \textbf{+reward\_norm}  & \textbf{Ours(Best)}  \\ \midrule
    Walker2d-m   & 80.1 ± {\scriptsize 1.4} & 80.1 ± {\scriptsize 0.8} (+0.0\%) & 84.8 ± {\scriptsize 2.3 } (+5.9\%) & \textbf{86.6} ± {\scriptsize 2.8} (+8.1\%)& 88.3\\
    Walker2d-m-r & 79.3 ± {\scriptsize 1.5} & \textbf{81.7} ± {\scriptsize 1.9} (+3.0\%)&  65.4 ± {\scriptsize 3.0 } (-17.5\%)  &  80.8 ± {\scriptsize 2.8} (+1.9\%) & 84.4  \\
    Walker2d-m-e & 109.5 ± {\scriptsize 0.6} & \textbf{109.7} ± {\scriptsize 0.3} (+0.2\%) &  108.4 ± {\scriptsize 0.8} (-1.0\%) & \textbf{109.7} ± {\scriptsize 0.3} (+0.2\%) & 111.1  \\
    Hopper-m     & 81.3 ± {\scriptsize 8.8} & \textbf{92.2} ± {\scriptsize 3.2} (+13.5\%)&  91.4 ± {\scriptsize 1.8} (+1.2\%) & 90.7 ± {\scriptsize 0.6} (+11.7\%)& 94.5 \\
    Hopper-m-r   & 84.5 ± {\scriptsize 12.8} & 88.2 ± {\scriptsize 2.3} (+4.4\%) &  \textbf{95.6} ± {\scriptsize 1.8} (+13.1\%)  & 92.5 ± {\scriptsize 0.9} (+9.5\%) & 96.0  \\
    Hopper-m-e   & 109.4 ± {\scriptsize 1.8} & 105.6 ± {\scriptsize 4.9} (-3.5\%) & 100.8  ± {\scriptsize 8.2} (-7.9\%) & \textbf{109.5} ± {\scriptsize 2.4} (+0.1\%)& 112.3\\
    Halfcheetah-m & 42.7 ± {\scriptsize 0.3}  & 40.8 ± {\scriptsize 0.6} (-4.4\%)&  43.0 ± 0.4 {\scriptsize } (+0.7\%) & \textbf{43.6} ± {\scriptsize 0.3} (+2.2\%)& 44.2 \\
    Halfcheetah-m-r & 39.9 ± {\scriptsize 0.7} & 40.5 ± {\scriptsize 0.5} (+1.4\%)&   39.6 ±  {\scriptsize 0.5} (-0.7\%) & \textbf{40.6} ± {\scriptsize 0.2} (+1.8\%)& 41.6 \\
    Halfcheetah-m-e & 78.2 ± {\scriptsize 11.6} & 83.9 ± {\scriptsize 6.4} (+7.3\%)&  87.6  ± {\scriptsize 1.6 } (+12.0\%) & \textbf{91.9} ± {\scriptsize 0.4} (+17.5\%)& 93.5 \\ \midrule
    \textbf{Average} & 78.32 ± {\scriptsize 4.36} & 80.30 ± {\scriptsize 2.32} (+2.5\%) & 79.63 ± {\scriptsize 2.27} (+1.7\%)  & \textbf{82.88} ± {\scriptsize 1.20} (+5.8\%)&  85.10  \\ \bottomrule
\end{tabular}
% \end{adjustbox}
\caption{Comparison of ablation performance with preference-based rewards. The results are trained with preference-based reward datasets, consistent with the average score recorded in Table \ref{tab:abl}.}
\label{tab:gt2}
\end{table}

\subsubsection{E6. Additional Experiments on Ensemble Normalization} \

The proposed ensemble normalization (EN) is a general mechanism, and we further demonstrate its effectiveness through application to Pb-IQL and Pb-TD3BC.
The results in Table \ref{tab:en_performance} demonstrate that EN provides performance improvements. However, DTR without EN still outperforms all baselines, highlighting the critical role of DT-based in-dataset regularization.

\begin{table}[h]
\centering
\label{tab:en_performance}
\begin{tabular}{lccccc}
\toprule
             & Pb-IQL & Pb-IQL + EN & Pb-TD3BC & Pb-TD3BC + EN & DTR w/o EN \\ \midrule
    Mujoco-Average         & 73.29  & 74.24 (+1.3\%) & 51.86   & 64.08 (+23.5\%) & 80.3       \\ \bottomrule
\end{tabular}
\caption{Performance Comparison with and without Ensemble Normalization}
\label{tab:en_performance}
\end{table}

\subsubsection{E7. Additional Experiments to Support Theoretical Findings} \

In this section, we present additional explanation and experimental results to further support the theoretical insights outlined in Equation \ref{eq:subopt}.

First, we explain how Theorem \ref{theorem1} differs from previous theoretical approaches. 
Our theorem extends offline RL theory specifically to offline PbRL, offering insights into how in-dataset trajectory can be beneficial. Unlike \cite{hu2023provable}, which assumes ground-truth and reward-free data, and \cite{provable24}, which assumes an optimal downstream reward in offline PbRL, we consider both reward learning bias (first term in Equation \ref{eq:subopt}) and errors in offline RL under limited data (second term in Equation \ref{eq:subopt}), leading to a theoretical upper bound for offline PbRL.

The upper bound underscores the importance of in-dataset trajectories. 
Specifically, the first term in Equation \ref{eq:subopt} demonstrates that increasing  $N_p$ (preference trajectories) reduces bias, leading to improved performance, as illustrated in Figure \ref{fig:abl_three3}. Meanwhile, the second term shows that increasing $N_o$ (in-dataset trajectories) reduces estimation error and further enhances performance. 
A supplementary experiment, conducted using only the preference dataset ($N_o=0$), reveals a significant drop in performance, highlighting the critical role of in-dataset offline data.
Similar results have been empirically demonstrated in the field of offline RL \cite{hu2023provable,zhang2024high}, showcasing the advantages of high-quality data in enhancing performance.

\begin{table}[h]
\centering
\label{tab:ablation}
\begin{tabular}{lcc}
\toprule
    & \textbf{DTR} & \textbf{DTR with }$N_o=0$ \\ \midrule
    walker2d-m    & 86.6 ± {\scriptsize 5.7}   & 82.4 ± {\scriptsize 0.7} (-4.8\%) \\
    walker2d-m-r  & 80.8 ± {\scriptsize 5.6}   & 73.2 ± {\scriptsize 5.4} (-9.4\%) \\
    walker2d-m-e  & 109.7 ± {\scriptsize 0.6}  & 93.4 ± {\scriptsize 16.9} (-14.9\%) \\
    hopper-m      & 90.7 ± {\scriptsize 1.3}   & 90.7 ± {\scriptsize 3.6} (-0.0\%) \\
    hopper-m-r    & 92.5 ± {\scriptsize 1.9}   & 49.0 ± {\scriptsize 19.4} (-47.0\%) \\
    hopper-m-e    & 109.5 ± {\scriptsize 4.7}  & 99.2 ± {\scriptsize 7.0} (-9.4\%) \\
    halfcheetah-m & 43.6 ± {\scriptsize 0.6}   & 43.0 ± {\scriptsize 0.1} (-1.4\%) \\
    halfcheetah-m-r & 40.6 ± {\scriptsize 0.5} & 40.2 ± {\scriptsize 0.2} (-1.1\%) \\
    halfcheetah-m-e & 91.9 ± {\scriptsize 0.7} & 89.8 ± {\scriptsize 1.1} (-2.3\%) \\ \midrule
    \textbf{Average} & \textbf{82.9 }± {\scriptsize 4.2} & 73.4 ± {\scriptsize 6.0} (-11.4\%) \\
    \bottomrule
\end{tabular}
\caption{Performance Comparison of DTR with and without In-Dataset Trajectories.}
\end{table}

\subsubsection{E8. Explanation of Possible Reasons Behind the Performance of HalfCheetah and Pen} \

Some attentive readers may have noticed that the performance score of DTR on HalfCheetah and Pen is lower than other methods across all datasets. 
Although it is challenging for any algorithm to achieve the best performance in every task, we provide some potential reasons for this phenomenon to allow readers to form a comprehensive evaluation of our method:

\begin{enumerate}
    \item Misalignment between GT rewards and human preferences: GT rewards may prioritize behaviors that are inconsistent with human intent, such as bipedal movement in HalfCheetah. This misalignment can lead to learned behaviors that better align with human preferences but result in lower scores according to the GT reward metrics.
    \item The inherent limitation of the transformer's stitching ability: As observed in Decision Transformer (DT) and subsequent studies, the performance of transformer-based methods like DTR on HalfCheetah tends to be weaker compared to traditional TD-learning-based methods. This highlights a possible deficiency in leveraging past trajectories effectively for complex control tasks.
\end{enumerate}

\end{document}